%% file: main.tex
\documentclass[10pt,twocolumn,letterpaper]{article}

\usepackage[preprint]{iccv}

\usepackage[utf8]{inputenc} % allow utf-8 input
\usepackage[T1]{fontenc}    % use 8-bit T1 fonts
\usepackage{url}            % simple URL typesetting
\usepackage{booktabs}       % professional-quality tables
\usepackage{amsfonts}       % blackboard math symbols
\usepackage{nicefrac}       % compact symbols for 1/2, etc.
\usepackage{microtype}      % microtypography
\usepackage{xcolor}         % colors
\usepackage{amsmath}
\usepackage{amssymb}
\usepackage{graphicx}
\usepackage{multirow}
\usepackage{xcolor}
\usepackage{colortbl}
\newcommand{\tablestyle}[2]{\setlength{\tabcolsep}{#1}\renewcommand{\arraystretch}{#2}\centering\footnotesize}
\newcommand{\modelname}{ViLLa}
\newcommand*{\Rom}[1]{\expandafter\@slowromancap\romannumeral #1@}

\definecolor{citecolor}{HTML}{0071bc}
\usepackage[colorlinks, linkcolor=red, colorlinks, anchorcolor=blue, citecolor=citecolor, pagebackref=True]{hyperref}
\def\paperID{10693} % *** Enter the Paper ID here
\def\confName{ICCV}
\def\confYear{2025}

\title{ViLLa: Video Reasoning Segmentation with Large Language Model}

\author{
\bf Rongkun Zheng$^{1}$ \quad Lu Qi$^{2}$ \quad Xi Chen$^{1}$ \quad
Yi Wang$^{3}$ \\ \bf \quad Kun Wang$^{4}$ \quad Yu Qiao$^{3}$ \quad Hengshuang Zhao$^{1}$\thanks{Corresponding author} \\
$^{1}$The University of Hong Kong $^{2}$Wuhan University \\ $^{3}$Shanghai Artificial Intelligence Laboratory $^{4}$SenseTime Research \\
\texttt{\{zrk22@connect, hszhao@cs\}.hku.hk}
}
\begin{document}

\maketitle

\begin{abstract}
Recent efforts in video reasoning segmentation (VRS) integrate large language models (LLMs) with perception models to localize and track objects via textual instructions, achieving barely satisfactory results in simple scenarios. However, they struggled to discriminate and deduce the objects from user queries in more real-world scenes featured by long durations, multiple objects, rapid motion, and heavy occlusions. In this work, we analyze the underlying causes of these limitations, and present \textbf{\modelname}: \textbf{Vi}deo reasoning segmentation with \textbf{L}arge \textbf{La}nguage Model. Remarkably, our ViLLa manages to tackle these challenges through multiple core innovations: (1) a context synthesizer that dynamically encodes the user intent with video contexts for accurate reasoning, resolving ambiguities in complex queries, and (2) a hierarchical temporal synchronizer that disentangles multi-object interactions across complex temporal scenarios by modelling multi-object interactions at local and global temporal scales. To enable efficient processing of long videos, {\modelname} incorporates (3) a key segment sampler that adaptively partitions long videos into shorter but semantically dense segments for less redundancy. What's more, to promote research in this unexplored area, we construct a VRS benchmark, \textit{VideoReasonSeg}, featuring different complex scenarios. Our model also exhibits impressive state-of-the-art results on VideoReasonSeg, Ref-YouTube-VOS, Ref-DAVIS17, MeViS, and ReVOS. Both quantitative and qualitative experiments demonstrate that our method effectively enhances video reasoning segmentation capabilities for multimodal LLMs. 
\end{abstract}

\input{Sec/1_Intro}
\input{Sec/2_Related}
\input{Sec/3_Method}
\input{Sec/5_Experiment}
\input{Sec/6_Conclusion}
\clearpage

\small
\bibliographystyle{plain}
\bibliography{main}

\clearpage

\newpage
\appendix
\section*{Appendix}

This supplementary material provides more details about the proposed ViLLa, and our proposed benchmark, VideoReasonSeg. The first part includes discussions about the design of ViLLa and its comparison with previous methods, followed by the implementation details. Then, we provide extra ablation experiments. What's more, we include the data generation pipeline of our VideoReasonSeg dataset, and further data cases demonstrating the variety of our data sample. Finally, we include failure case analysis. The content is organized as follows:
\begin{itemize}
\setlength{\itemsep}{1pt}
\setlength{\parsep}{1pt}
\setlength{\parskip}{1pt}
\item {Discussions of ViLLa's differences with previous methods.}
\item{The implementation details of ViLLa.}
\item{More ablation study experiment of ViLLa.}
\item{The data details and the data generation pipeline of our proposed VideoReasonSeg dataset.}
\item{Failure case analysis of ViLLa.}
\end{itemize}

\vspace{-5pt}
\section{Discussions}
ViLLa is a holistically designed framework for Video Reasoning Segmentation (VRS), a burgeoning task demanding spatiotemporal tracking, segmentation, and dynamic reasoning in complex, evolving scenes. While prior works (e.g., LISA, PixelLM, Mask2Former) focus on reasoning static images or tracking through short clips, their architectures inherently mismatch VRS needs for two major reasons: \textbf{1) Temporal Reasoning Gap:} Static MLLMs lack mechanisms to model motion, causality, or long-term dependencies in videos, while VOS/VIS methods lacks high-level reasoning beyond tracking. \textbf{2) Integration Challenges:} Simply grafting spatial reasoning modules (e.g., ViLLa†) onto temporal trackers incurs additional computational costs and suboptimal performance. ViLLa bridges these barriers via three novel, task-specific components: \romannumeral1) Key Segment Extractor: Identifies critical temporal segments to reduce redundancy. \romannumeral2) Context Synthesizer: Fuses and condenses long-term spatial-semantic reasoning information with dynamic scene evolution. \romannumeral3) Hierarchical Temporal Synchronizer: Ensures consistency across long-term dependencies and modeling of complex scenes via multi-scale aggregation with multi-level segmentation tokens. As shown in Tab.~\ref{tab:adaptation}, ViLLa achieves +3.5/2.4 gains over ViLLa† (adapted from LISA with designs from PixelLM, Mask2Former, and curated data), proving that adaptive integration of spatiotemporal reasoning modules—not direct reuse of static models—is essential and effective for VRS. In short, our VRS-oriented design novelly addresses understudied challenges in VRS.

\begin{table}[h]
    \centering
    \scriptsize
    \caption{\textbf{Comparison} of direct adaptation of former approaches.}
    \vspace{-5pt}
    \resizebox{1.0\linewidth}{!}
    {
        \setlength{\tabcolsep}{10pt}
        \begin{tabular}{l | c c | c c c}
        \toprule
        \multirow{2}*{Method} &  \multicolumn{2}{c |}{VideoReasonSeg} & \multicolumn{3}{c}{MeViS} \\
          & \( \mathcal{J} \)\&\( \mathcal{F} \) & Accuracy   &  \( \mathcal{J} \)\&\( \mathcal{F} \) & \( \mathcal{J} \) & \( \mathcal{F} \) \\
        \midrule
        ViLLa$^\dag$  &51.9  &44.0  &47.0  &43.9  &50.1  \\
        \cellcolor[HTML]{efefef}\textbf{ViLLa} &\cellcolor[HTML]{efefef}\textbf{55.4} &\cellcolor[HTML]{efefef}\textbf{49.9}  &\cellcolor[HTML]{efefef}\textbf{49.4} &\cellcolor[HTML]{efefef}\textbf{46.5} &\cellcolor[HTML]{efefef}\textbf{52.3}\\
        \bottomrule 
        \end{tabular}
    }
    \label{tab:adaptation}
    \vspace{-0.1in}
\end{table}

\begin{table}[t]
    \centering
    \scriptsize
    \caption{
        \textbf{Training hyperparameters} for ViLLa.
    }\label{tab:hyperparameters} 
    \vspace{-5pt}
    \resizebox{0.84\linewidth}{!}{
        \tablestyle{15pt}{1.0}
        \begin{tabular}{l|c}
        \toprule
        Config & Value \\
        \midrule
        input resolution & 224 \\
        max text length  & 512 \\
        optimizer & {AdamW} \\ 
        optimizer momentum & $\beta_1$, $\beta_2$ = 0.9, 0.999 \\
        weight decay & {0.02} \\
        learning rate schedule & {cosine decay} \\
        learning rate & 2e-5 \\
        batch size & 32 \\
        warmup iters & 10 \\
        % $\lambda_{txt}$ & 1.0 \\
        % $\lambda_{mask}$ & 1.0 \\
        % $\lambda_{dice}$ & 0.5 \\
        % $\lambda_{ce}$ & 2.0\\
        \bottomrule
        \end{tabular}
    }
\vspace{-10pt}
\end{table}

\vspace{-5pt}
\section{Implementation Details}
\noindent\textbf{Training Details.}
In the first part, we present the detailed training configuration in our Tab.~\ref{tab:hyperparameters}. As for the $\lambda_{txt}$ and $\lambda_{mask}$, they are set to 1.0, and $\lambda_{dice}$ is 0.5 while $\lambda_{ce}$ is 2.0.

\vspace{-10pt}
\section{Additional Ablation Experiments}
\noindent \textbf{Sampling Strategy.}
In the key segment extractor, we take the average from the top-K responses to obtain the starting and ending frames of the key segments, where we denote as $V_{key}$, comprising $T_{key}$ frames. Based on the key segments, we also sample $T_{ref}$ using an adaptive global sampling strategy. `Global' indicates sampling $T_{ref}$ frames from the whole video apart from the key segments we extract, and `neighbor' denotes sampling frames from the neighboring frames (both precedent and antecedent) of the key segments. Our `adaptive' sampling strategy, on the other hand, combines both `global' and `local' sampling strategies, which samples $T_{ref}/3$ from the whole video, and $2/3 T_{ref}$ from the neighboring frames. As shown in Tab.~\ref{tab:sampling}, adaptive sampling slightly outperforms both global and neighbor sampling strategies. As the number $T_{ref}$ increases, the performance gradually improves.

\noindent \textbf{Aggregation Strategy.}
Tab.~\ref{tab:aggregation} shows the results of different aggregation strategies in the segment synchronization decoder. We compare our aggregation between video-level segmentation embeddings with the feature fusion adopted in PixelLM~\cite{ren2023pixellm}. As shown in the table, our strategy improves the performance on referring VOS dataset, demonstrating the effectiveness of the video-frame aggregation strategy.

\noindent
\textbf{Video Instance Segmentation.}
Tab.~\ref{tab:vis} presents the results on the video instance segmentation datasets. YouTube-VIS 2019~\cite{yang2019video}, contains 2.9k videos. The dataset was updated to YouTube-VIS 2021 with longer videos. OVIS dataset is another resource for video instance segmentation, particularly focusing on scenarios with severe occlusions between objects~\cite{qi2022occluded}. It consists of 25 object categories and 607 training videos. Our ViLLa surpasses previous SOTA VIS methods by 3.3, 3.8, and 3.9 points, respectively. The results prove that our model is excelling at modeling temporal relations and segmenting high-quality tracklets. 

\input{Tab/sampling}

\begin{table}[t]
    \centering
    \caption{\textbf{Ablation study} on aggregation strategies.}
    \label{tab:aggregation}
    \scriptsize
    \vspace{-5pt}
    \resizebox{1.0\linewidth}{!}
    {
        \begin{tabular}{l | c c c | c c c}
        \toprule
        \multirow{2}*{Strategy} &  \multicolumn{3}{c |}{Ref-YouTube-VOS} & \multicolumn{3}{c}{Ref-DAVIS17} \\
          & \( \mathcal{J} \)\&\( \mathcal{F} \) & \( \mathcal{J} \) & \( \mathcal{F} \)  &  \( \mathcal{J} \)\&\( \mathcal{F} \) & \( \mathcal{J} \) & \( \mathcal{F} \) \\
        \midrule
        Feature Fusion  &65.9  &64.1  &68.5  &63.2  &60.5  &66.1  \\
        \textbf{Embedding Similarity} &\textbf{66.5}& \textbf{64.6}& \textbf{68.6}&\textbf{64.4} &\textbf{61.2} &\textbf{67.7}\\
        \bottomrule 
        \end{tabular}
    }
\vspace{-5pt}
\end{table}

\begin{table}[t]
    \centering        
    \caption{\textbf{Comparison} on multiple VIS datasets.}
    \label{tab:vis}
    \vspace{-5pt}
    \resizebox{\linewidth}{!}{
        \tablestyle{12pt}{1.0}
        \begin{tabular}{l|c|c|c}
        \toprule
        \multirow{2}*{Method} &YTVIS-19 &YTVIS-21 &OVIS \\
        &AP &AP &AP\\
        \midrule
        SeqFormer~\cite{wu2022seqformer} &59.3 &51.8 &- \\
        Mask2Former~\cite{cheng2021mask2former} &61.6 &55.3 &24.1 \\
        VITA~\cite{heo2022vita} &63.0 &57.5 &27.7 \\
        IDOL~\cite{wu2022defense} &64.3 &56.1 &42.6 \\
        \cellcolor[HTML]{efefef}{\textbf{ViLLa}} &\cellcolor[HTML]{efefef}\textbf{67.6} &\cellcolor[HTML]{efefef}\textbf{59.9} &\cellcolor[HTML]{efefef}\textbf{46.5} \\
        \bottomrule
        \end{tabular}
    }
\vspace{-15pt}
\end{table}

\vspace{-5pt}
\section{VideoReasonSeg Details}
%\noindent\textbf{Data Generation Pipeline.}
In order to generate multiple-choice QA, we automatically convert the video annotations into this format via LLMs. Specifically, we first use ChatGPT~\cite{chatgpt} to generate a question for each video. For most questions, we construct the option candidates directly from the ground truth annotations. For example, video segmentation tasks contain masks and instance categories of each video. Then the candidate option for multiple-choices would be the \textit{correct} category, \textit{wrong} category, and a \textit{not-sure} choice. Ultimately, we produce 2 pairs for each of the video. To strengthen the evaluation's robustness, for each question we randomly sample 3 to 5 answer options from the available candidates and shuffle the order of the options. Additionally, to prevent the common issue of answer leakage where longer options tend to be correct, we further use a large language model to ensure that all the answer options for a question are of similar and reasonable lengths.

\begin{figure*}[h]
    \centering
    \includegraphics[width=1.0\textwidth]{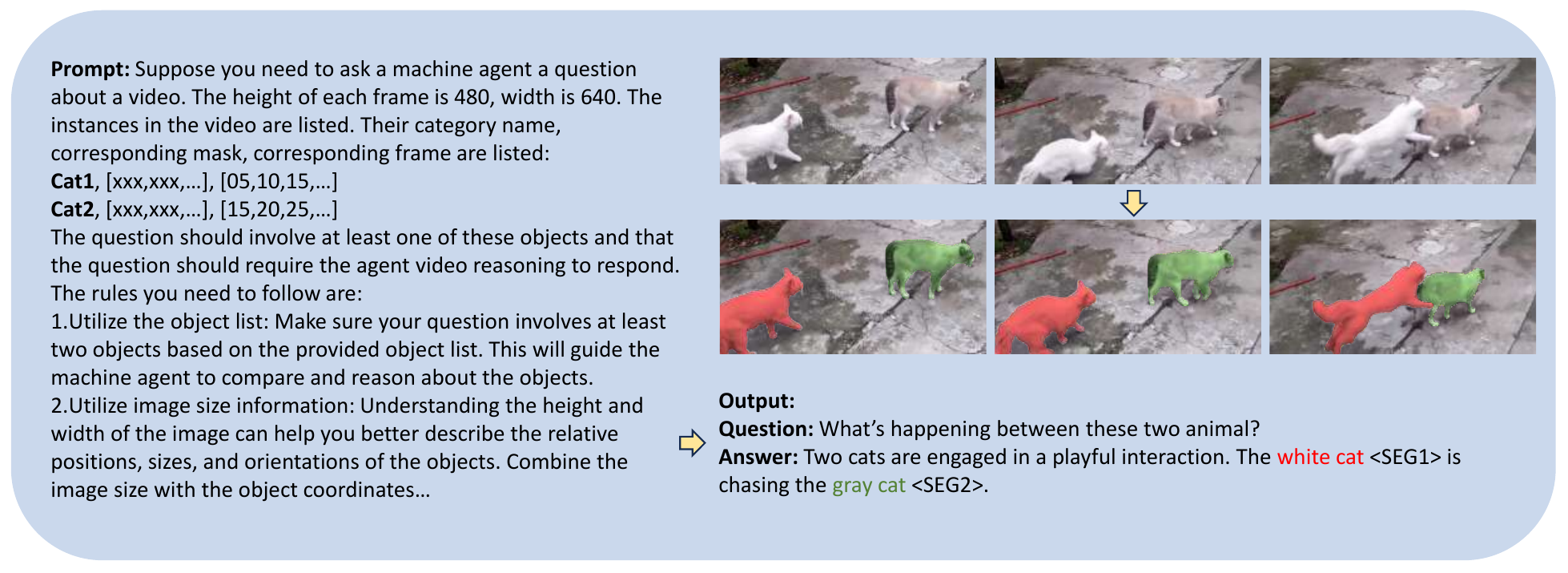}
    \caption{{\textbf{GPT-4V data generation pipeline.} The right part shows an example of how reasoning segmentation data and multiple choices are generated. The input prompt includes certain rules and the position as well as time localizations to instruct GPT-4V into generating more effective data samples.}}
    \label{fig:data}
    %\vspace{-10pt}
\end{figure*}

\begin{figure*}[t]
    \centering
    \includegraphics[width=0.95\textwidth]{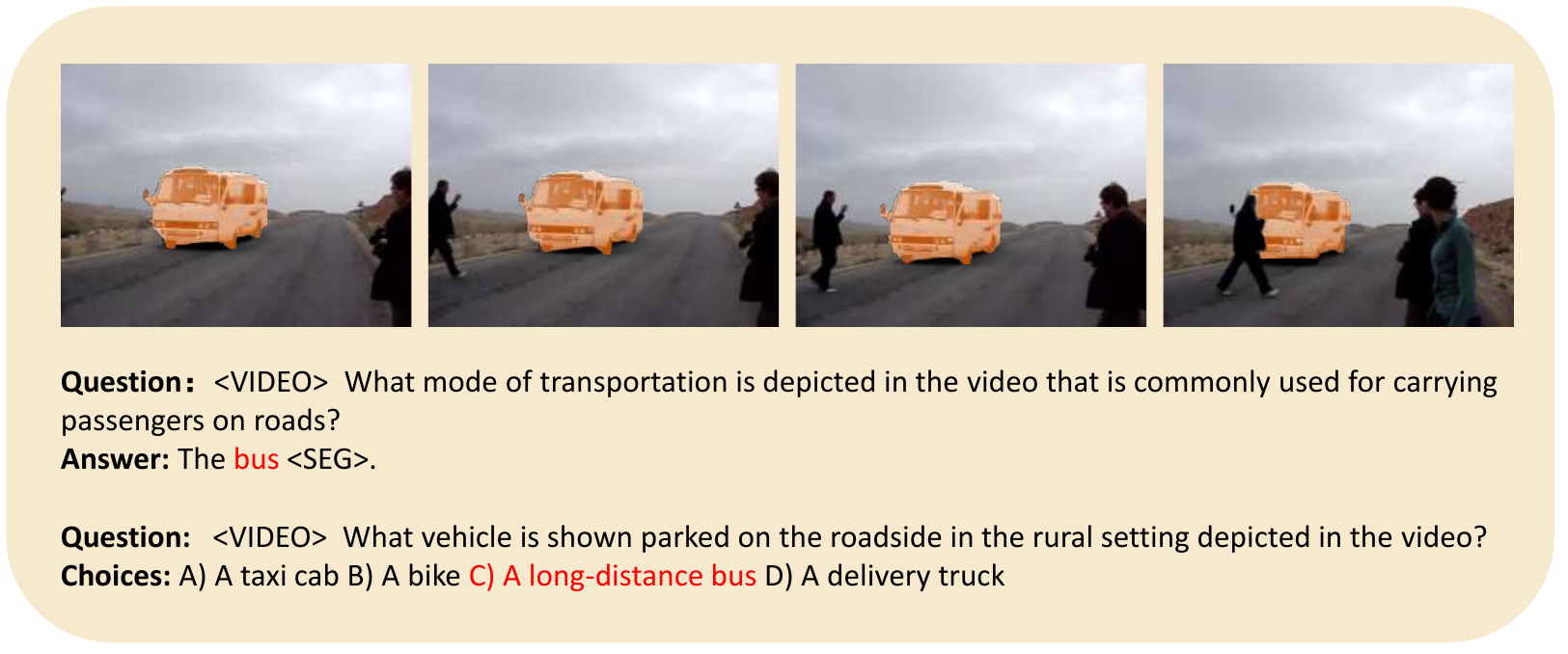}
    \caption{{\textbf{GPT-4V data generation samples.} The part shows further samples of the generated questions and the multiple choices. The two types of questions can help us better evaluate the model's performance in video reasoning at both pixel-level and video-level.}}
    \label{fig:data_example}
    %\vspace{-10pt}
\end{figure*}

As for the question and answer pair, we use GPT-4V to construct our dataset. We utilize videos with pre-existing video mask annotations. The video frames, the category names contained in the video, and their related mask annotation are contained in the prompts fed to the GPT-4V. An example of the prompts is shown in Fig.~\ref{fig:data}. Using carefully crafted prompts, GPT-4V autonomously selects instances to construct question-answer pairs relevant to the video. As illustrated in Fig.~\ref{fig:data_example}, we demonstrate two types of questions, both question and multiple choice, from a given video. In this example, we show that our data tests the capability of models of reasoning based on common world-knowledge, and relate `vehicle carrying passengers' to the white bus on the roadside. In addition, the multiple choice expects the model to distinguish the type of vehicle from other plausible answers, such as `taxi' and `bike'. All together these questions are tests of the reasoning capacities of models on both pixel-level and video-level.

Although GPT-4V can efficiently understand the content of the video frame, there are still failure cases in the generated data. One major problem is that questions can be too objective and hard to evaluate. For example, the question ``How would you rate the overall difficulty and impressiveness of the skateboarding you observed?'' is very objective, and the answers can vary for different people. This requires further prompts and filtering during data generation process. 

\begin{figure*}[t]
	\centering
	\small
	\resizebox{1.0\linewidth}{!}{
	\begin{tabular}{m{0.01cm}m{14cm}}
		% Input & Ground Truth & Point Transformer & Input & Ground Truth & Point Transformer \\
		{(a)} &\includegraphics[width=\linewidth]{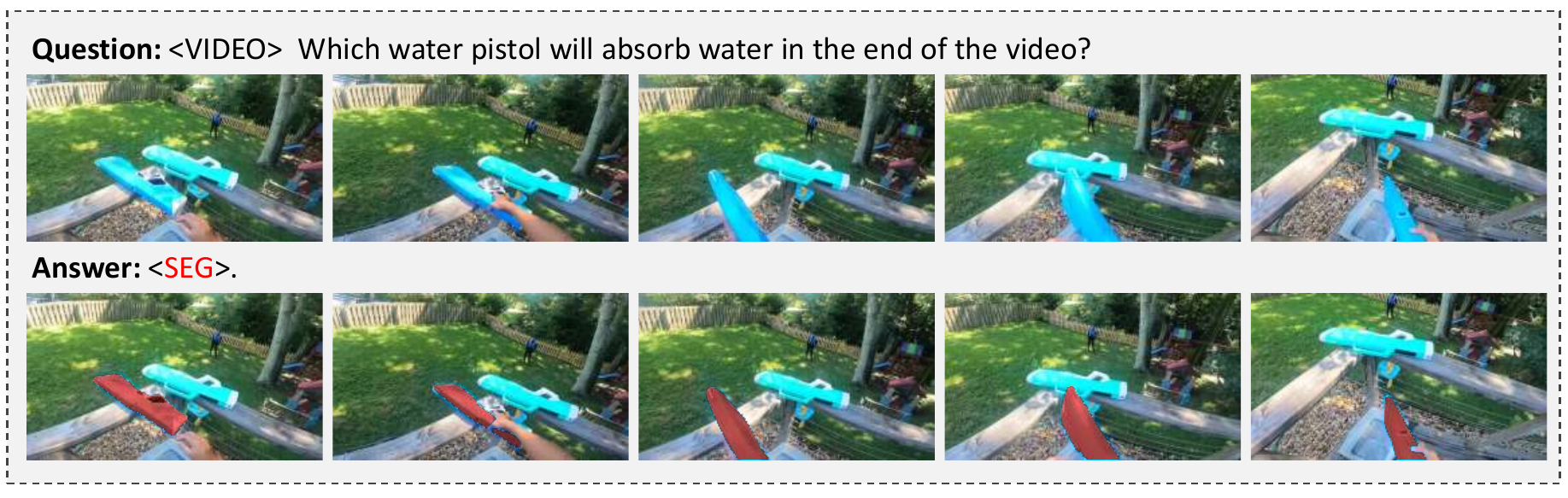}\\
		{(b)} &\includegraphics[width=\linewidth]{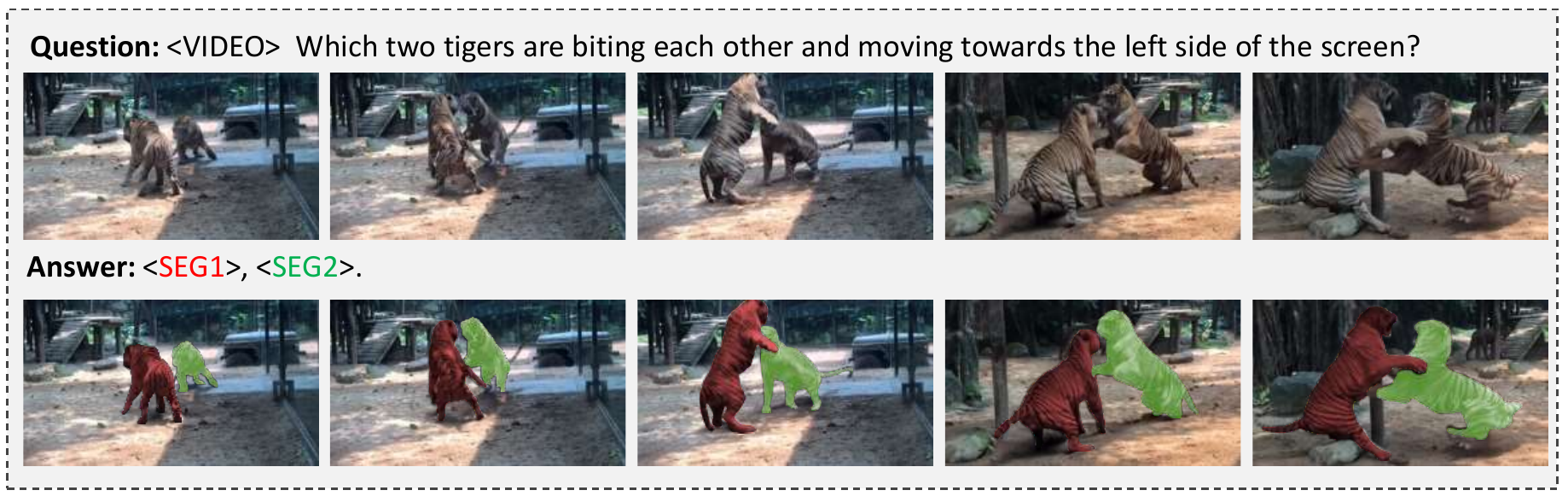}\\
        {(c)} &\includegraphics[width=\linewidth]{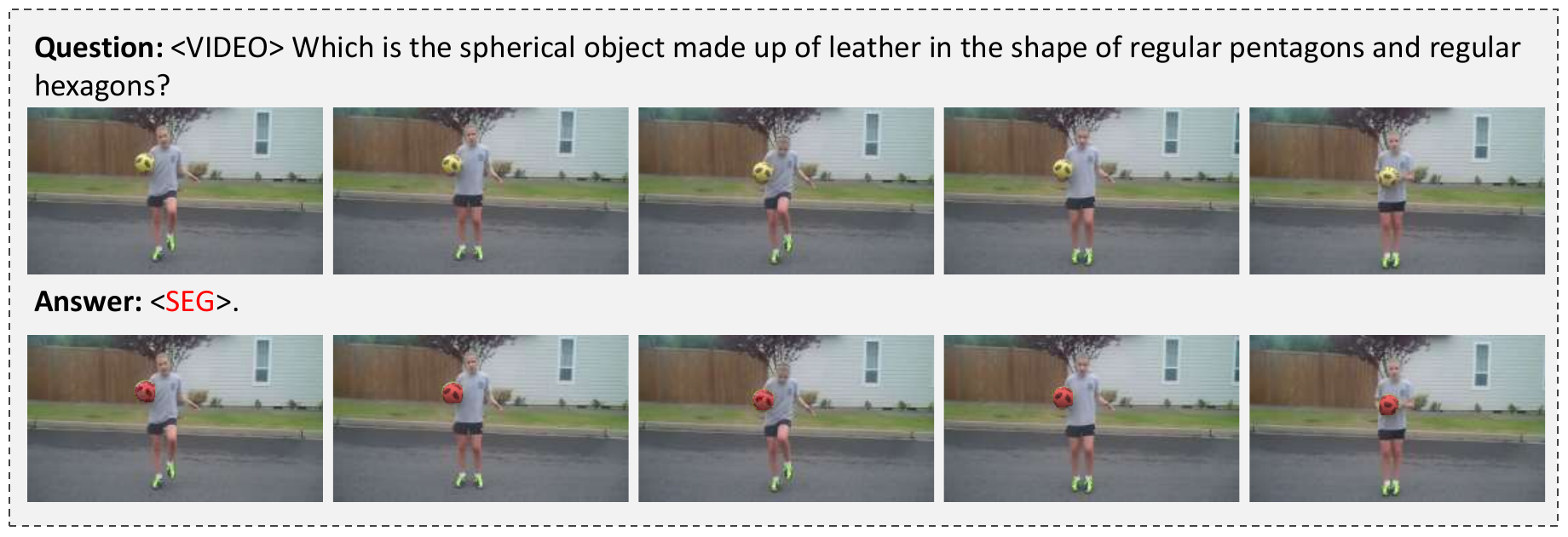}\\
	\end{tabular}}
        %\vspace{-10pt}
	\caption{\textbf{Data samples} of our proposed VideoReasonSeg.}
        %\vspace{-10pt}
	\label{fig:supp_vis}
\end{figure*}

Video visualizations of additional data of the VideoReasonSeg are further presented in Fig.~\ref{fig:supp_vis}, which shows varied cases that include: a) discrimination from multiple instances; b) multiple instances with fast movement; c) open-world knowledge reasoning.

\vspace{-5pt}
\section{Failure Case Analysis}
\begin{figure*}[h]
    \centering
    %\vspace{-5pt}
    \includegraphics[width=0.95\linewidth]{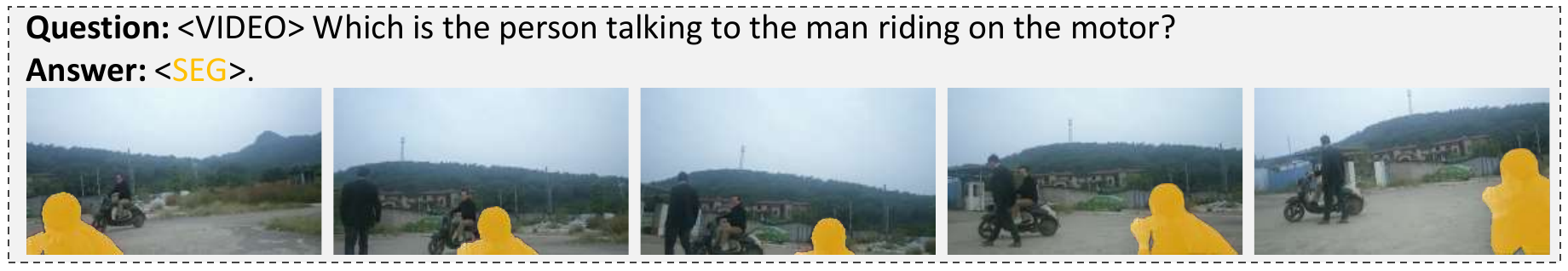}
    \caption{\textbf{Failure case.} ViLLa incorrectly segments the bystander who is observing the two-person talking.}
    \label{fig:model1}
\end{figure*}

\vspace{-5pt}
 Even though our ViLLa shows impressive results in video reasoning segmentation, there is still room for improvement. As shown in Fig.~\ref{fig:model1}, ViLLa incorrectly segments the bystander who is observing the two-person talking. We hypothesize that this error arises from the inability of the MLLM to temporally localize the ``talking'' action, which occurs exclusively in the final three frames (there is even an occlusion in the last frame). Consequently, ViLLa erroneously associates the individual co-occurring with the motor-riding man in the initial frames and persistently tracks this subject throughout the whole video.

%\clearpage

% {
% \small
% \bibliographystyle{ieee_fullname}
% %\bibliographystyle{plain}
% \bibliography{main}
% }

\end{document}

%% file: Sec/1_Intro.tex
\section{Introduction}
Capitalizing on the achievements of Large Language Models (LLMs)~\cite{vicuna,chatgpt,openai2023gpt4,zhang2023llama}, the development of large multimodal models (LMMs)~\cite{Qwen-VL,li2023blip,liu2023llava,2023interngpt,wang2023internvid,zhu2023minigpt} has notably enhanced visual perception capabilities and user-interaction experiences to new heights. However, in the absence of explicit instructions specifying target objects or categories, the majority of models~\cite{alayrac2022flamingo,li2023otter,li2023blip,liu2023improved,liu2023llava,ye2023mplug,zhu2023minigpt}, including both perception models and language ones, struggle to generate detailed and granular outputs such as instance masks. Instead, these models typically generate only general textual descriptions for images or videos. This limitation hinders the practical application of multimodal systems in various scenarios, such as autonomous driving~\cite{wang2024omnidrive}, image and video editing~\cite{liu2024referring}, robotics~\cite{sermanet2024robovqa}, and augmented reality~\cite{sadeghzadeh2024arva}.
%This limitation hinders the practical application of multimodal systems in industrial and everyday scenarios, such as autonomous driving~\cite{wang2024omnidrive}, image and video editing~\cite{liu2024referring}, robotics~\cite{sermanet2024robovqa}, and augmented reality~\cite{sadeghzadeh2024arva}.

% Video reasoning segmentation (VRS) is an emerging vision-language task that bridges semantic comprehension of textual instructions with pixel-level spatiotemporal localization in videos. Unlike conventional video segmentation—which focuses on low-level appearance or motion cues—VRS requires models to proactively interpret user intent (e.g., “Track the woman who handed a package to the driver”) and dynamically reason over complex interactions across frames. This capability is pivotal for real-world applications such as autonomous systems interpreting ambiguous navigation commands, surveillance systems analyzing long-term object behaviors, or assistive robots executing context-aware actions.

Recent studies~\cite{lai2023lisa,ren2023pixellm,yang2023improved} have investigated the application of LLMs to generate object masks in a novel reasoning segmentation task that requires complex reasoning to interpret sophisticated instructions. These approaches introduce efficient language models capable of producing multiple open-set targets while accommodating diverse reasoning complexities. However, despite their successes, these methods remain limited to image-level reasoning and do not extend fine-grained reasoning capabilities to the temporal dimension. To address this issue, researchers~\cite{yan2024visa,bai2024one} managed to equip the image reasoning segmentation model with temporal understanding ability, and propose VRS. VISA~\cite{yan2024visa} employs a video understanding model~\cite{li2025llama} to select the key frame for LISA~\cite{lai2023lisa} to segment and subsequently utilizes an object tracker~\cite{cheng2022xmem} to maintain temporal consistency. However, the quality of segmentation is highly dependent on the accuracy of key frame selection, which is susceptible to error accumulation through propagation as the number of input frames increases. On the other hand, VideoLISA~\cite{bai2024one} adopts a sparse dense sampling strategy to compress visual features and extends the single segmentation token to track an instance across frames, especially when the movement is rapid. Nevertheless, the sampling strategy fails to adaptively select the key video segments, making it challenging to capture the essential information from the video. Also, because a single segmentation token is unable to segment and track multiple instances across frames, the segmentation and tracking quality decrease significantly when the video scene becomes intricate (\textit{e.g.}, involving multiple objects with occlusion), as proven in various previous Video Instance Segmentation (VIS) approaches~\cite{heo2022vita,zhang2023dvis}, which utilize multiple-level of tokens to segment and track instances. 

% \begin{figure*}[t]
%     \centering
%     \includegraphics[width=0.85\linewidth]{Fig/teaser_1.pdf}
%     \caption{{Our ViLLa is an effective and efficient LLM capable of segmenting and tracking multiple objects in long videos while understanding the user's implicit instructions.}}
%     \label{fig:teaser}
%     \vspace{-5mm}
% \end{figure*}

\begin{figure*}[t]
	\centering
	\small
	\resizebox{0.96\linewidth}{!}{
	\begin{tabular}{m{0.01cm}m{14cm}}
		% Input & Ground Truth & Point Transformer & Input & Ground Truth & Point Transformer \\
		{(a)} &\includegraphics[width=\linewidth]{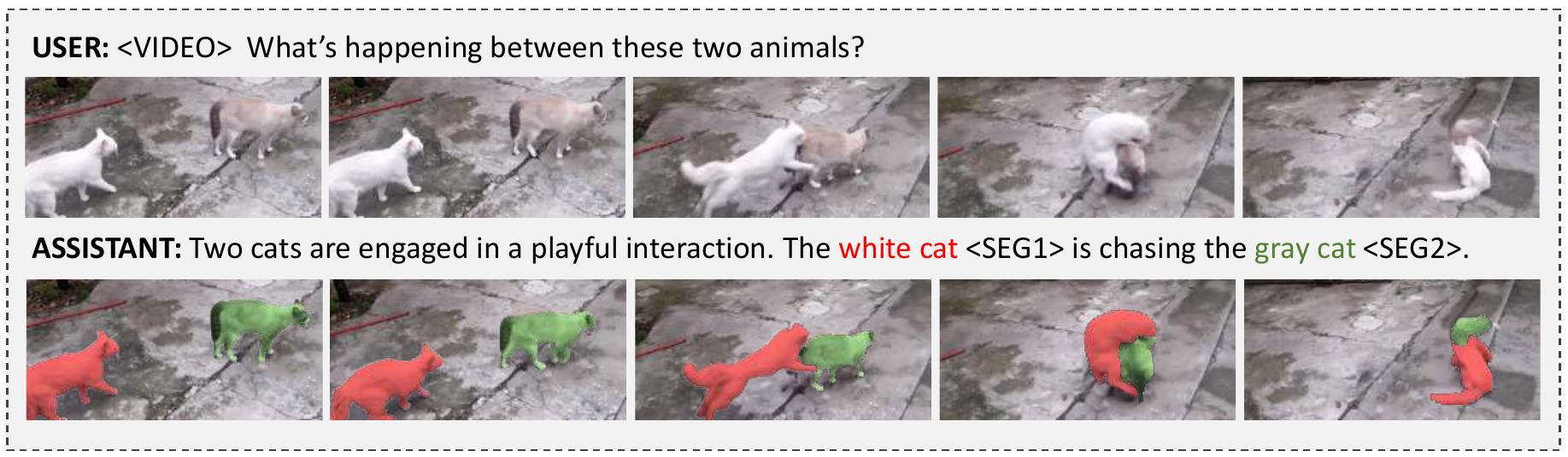}\\
		{(b)} &\includegraphics[width=\linewidth]{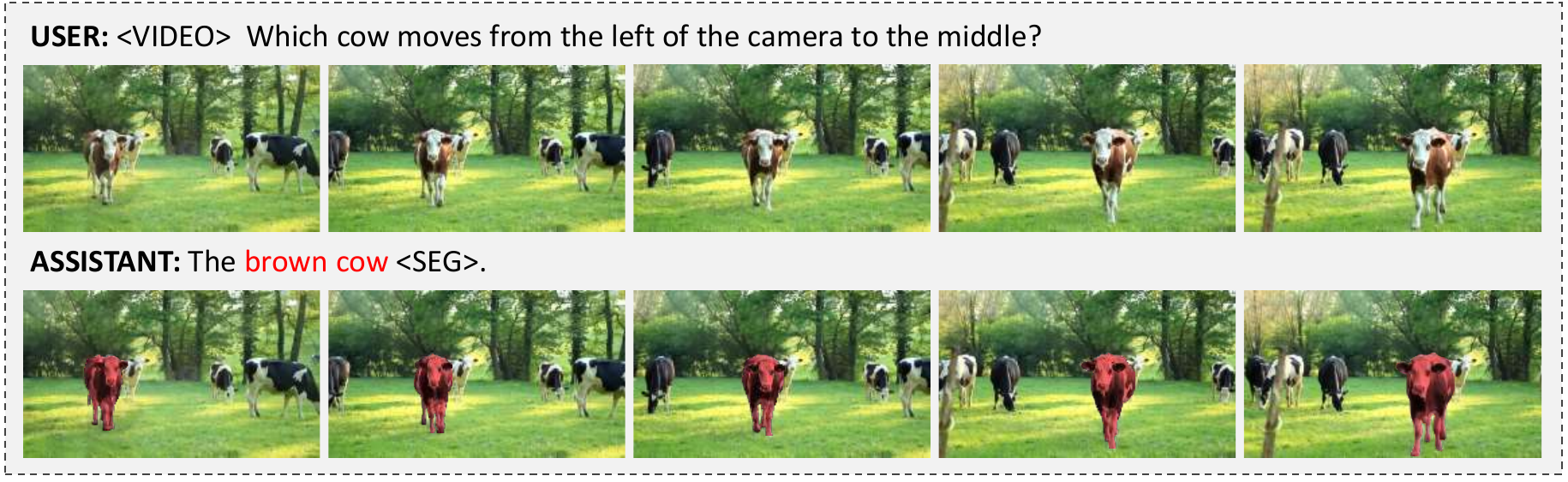}\\
        {(c)} &\includegraphics[width=\linewidth]{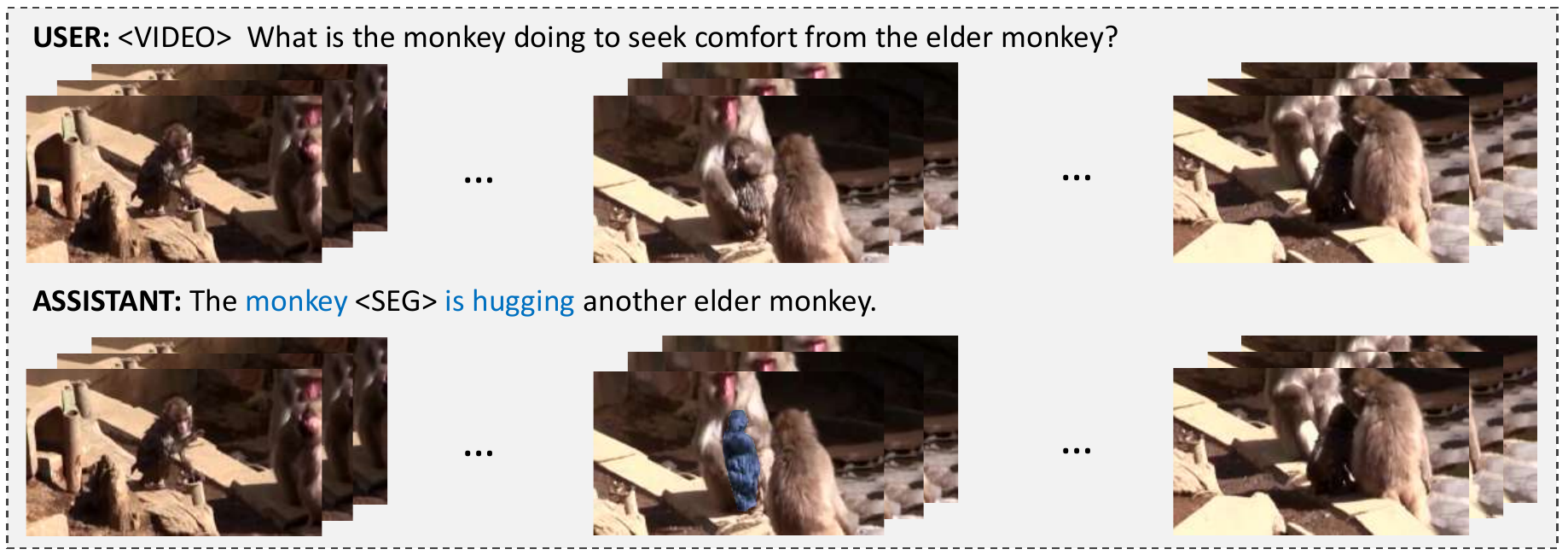}\\
	\end{tabular}}
        %\vspace{-10pt}
	\caption{Our ViLLa is an effective and efficient LMM capable of segmenting and tracking: (a) multiple objects with rapid motion; (b) objects in crowded scenes; (c) objects in long videos with occlusions.}
        \vspace{-10pt}
	\label{fig:teaser}
\end{figure*}

% Inspired by the exceptional capacity of LLMs to reason and comprehend user intentions, we aim to leverage this capability of LLMs and extend it from image to video. 
To tackle the aforementioned challenges, we propose \textbf{ViLLa}, an effective and efficient LMM for video reasoning segmentation. {\modelname} proficiently perceives and reasons videos with multiple objects, rapid motion, and diverse reasoning complexities (as shown in Fig.~\ref{fig:teaser}). Specifically, our {\modelname} has three key components: Key Segment Extractor, Context Synthesizer, and Hierarchical Temporal Synchronizer. These designs put effort into solving the complex video reasoning segmentation from different perspectives. The Key Segment Extractor selects the most relevant video segments, which reduces redundant visual tokens and effectively turns long videos into shorter segments. To make the responses more pertinent to users' intentions, Context Synthesizer synthesizes frame context information from visual features to text embeddings to generate contextually enriched text and highly input-responsive visual embeddings. To build up stronger temporal relations to multi-target segmentation tokens in complex scenes, the Hierarchical Temporal Synchronizer communicates and consolidates the multi-scale segmentation embeddings across temporal dimension via attention-based fusion strategy. Tested on multiple referring video object segmentation benchmarks, our model could surpass previous state-of-the-art LLM-based methods by a large margin. In addition, to validate its effectiveness, we establish a benchmark for video reasoning segmentation evaluation, called VideoReasonSeg. Utilizing a GPT-4V~\cite{openai2023gpt4}-aided data curation pipeline, we curate 3k high quality video samples for video reasoning segmentation spanning multiple-choice QAs and video-instruction-mask samples, and the video scenarios contain diverse reasoning complexities (as presented in the appendix), which offers convincing metrics. 

In summary, our contributions are as follows:
\begin{itemize}
    \item We analyze the limitations of current VRS approaches and propose a framework named ViLLa that segments and tracks the desired objects in more complex scenarios.
    \item We develop three key modules that resolves complex scenario reasoning from different perspectives, namely key segment extractor, context synthesizer, and hierarchical temporal synchronizer. Our extractor selects the most relevant video segments from the complex input video, while the synthesizer combines and infuses user intention and context information, and the synchronizer communicates multi-scale segmentation tokens spatio-temporally.
    \item We develop VideoReasonSeg, a comprehensive benchmark for VRS that spans different complex reasoning scenarios, comprising 3k videos and 15k object descriptions for instruction tuning and evaluation. This benchmark plays a crucial role in evaluating and fostering the exploration of reasoning capabilities in video-based models. % by the research community.
    \item We conduct extensive experiments on challenging referring segmentation benchmarks, ReVOS, and our proposed VideoReasonSeg. The state-of-the-art results demonstrate the effectiveness of our method.
\end{itemize}

%% file: Sec/2_Related.tex
\section{Related Works}
\noindent\textbf{Large Multimodal Models.}
Large multimodal models (LMMs) can be broadly categorized into two groups based on their utilization of large language models (LLMs). The first category comprises models~\cite{lu2022unified,wang2022ofa,yu2022coca} that are either trained from scratch or leverage smaller language models like BERT for text processing. These models typically employ a combination of contrastive and generative objectives to tackle a range of multimodal tasks (e.g., Coca~\cite{yu2022coca}). However, their limited language understanding capacity often restricts their performance in tasks that demand massive commonsense reasoning abilities.

The emergence of LLMs in the recent few years has paved the way for a new paradigm in LMM development, in which LLMs are augmented with multimodal comprehension abilities~\cite{zhang2022opt,touvron2023llama,alpaca,peng2023instruction}. This approach typically includes utilizing adapters to align visual and textual representations within LLMs, as exemplified by models such as Flamingo~\cite{alayrac2022flamingo}, BLIP-2~\cite{li2023blip}, MiniGPT-4~\cite{zhu2023minigpt}, LLaMA-Adapter~\cite{llama-adapter-v2}, LLaVA~\cite{liu2023llava}, InstructBLIP~\cite{instructblip}, InternGPT~\cite{2023interngpt}, QwenVL~\cite{Qwen-VL}, InternVideo2~\cite{wang2024internvideo2}. Some video-related MLLM (VideoLLM) VideoChatGPT~\cite{maaz2023video}, and Valley~\cite{luo2023valley} utilize ChatGPT~\cite{chatgpt} to generate video instruction-tuning data, aiming to enhance instruction-following capabilities for real-world video comprehension. Even though these models have demonstrated improved performance in vision-language tasks through instructional tuning, their primary limitation lies in generating only textual outputs.

\noindent\textbf{Video Segmentation.}
Video Instance Segmentation (VIS)~\cite{yang2019video} aims to detect, segment, and track object instances inside videos based on a set of predefined object categories simultaneously. Numerous studies~\cite{lin2021video,wu2022efficient,wang2021end,OMGSeg,dvisdaq} have proposed diverse designs (such as tracking by detection~\cite{athar2020stem}, masked attention~\cite{cheng2022masked}, object token association~\cite{wu2022seqformer,heo2022vita,li2022videoknet}, contrastive memory~\cite{heo2022generalized,ying2023ctvis}, and more) to effectively model temporal relationships. For example, IDOL~\cite{wu2022defense}, which is built upon Deformable-DETR~\cite{zhu2020deformable}, incorporates a contrastive learning head to obtain unique embeddings for associations~\cite{he2020momentum}. DVIS~\cite{zhang2023dvis}, on the other hand, proposed a decoupling strategy by dividing VIS into image segmentation, tracking, and refinement.
The Referring Video Object Segmentation (RVOS) task was introduced to segment the target instance referred by a given text in given videos. Early works~\cite{khoreva2019video,seo2020urvos} used the spatial-temporal memory mechanism to strengthen temporal referring relations. The researchers then started building offline reference models~\cite{wang2019asymmetric,wang2020context,wu2022multi,yan2024referred} that take the whole clip as input. Currently, query-based Transformer models~\cite{botach2022end,wu2022language} begin to dominate. MTTR~\cite{botach2022end} utilized an instance-level segmentation pipeline to determine the most suitable sequence corresponding to the referred object. ReferFormer~\cite{wu2022language} transformed text expressions into queries to attend to relevant regions in videos.

\noindent\textbf{Reasoning Segmentation.}
In numerous practical applications, it is essential to comprehend visual inputs at a more fine-grained level, such as specific regions. Kosmos-2~\cite{peng2023kosmos}, InternGPT~\cite{2023interngpt}, and Ferret~\cite{you2023ferret} offer grounding capabilities to specified image regions. Yet, these approaches still fail to provide pixel-level outputs. To bridge this gap, LISA~\cite{lai2023lisa} integrates SAM~\cite{kirillov2023segment} with LLMs to tackle segmentation tasks. PixelLM~\cite{ren2023pixellm} proposes a novel pixel decoder and a segmentation codebook to generate masks for multiple targets. Then, GLaMM~\cite{rasheed2024glamm} proposes region-level caption and segmentation tasks. Several works~\cite{hao2024vitron,liu2023llavaplus,qi2024generalizable} use LLMs as agents to assign different visual experts. OMG-LLaVA~\cite{zhang2024omg} implements a new comprehensive baseline that has only a single visual encoder. However, these methods mainly focus on reasoning image segmentation and fail to reason across temporal dimensions. When they are transferred to the video domain, the segmentation token is not capable of modeling multiple targets as well as their moving trajectories. Early approaches, namely VISA~\cite{yan2024visa} and VideoLISA~\cite{fu2024video} manage to bridge the gap by implementing different frame sampling strategies, but VISA suffers from the error accumulation from key frame selection and VideoLISA fails to adaptively extract relevant frames which leads to redundancy. After analyzing their drawbacks, we propose ViLLa with key segment extractor, context synthesizer, and hierarchical temporal synchronizer to address the problem.

\begin{figure*}[t]
\centering
\includegraphics[width=1.0\linewidth]{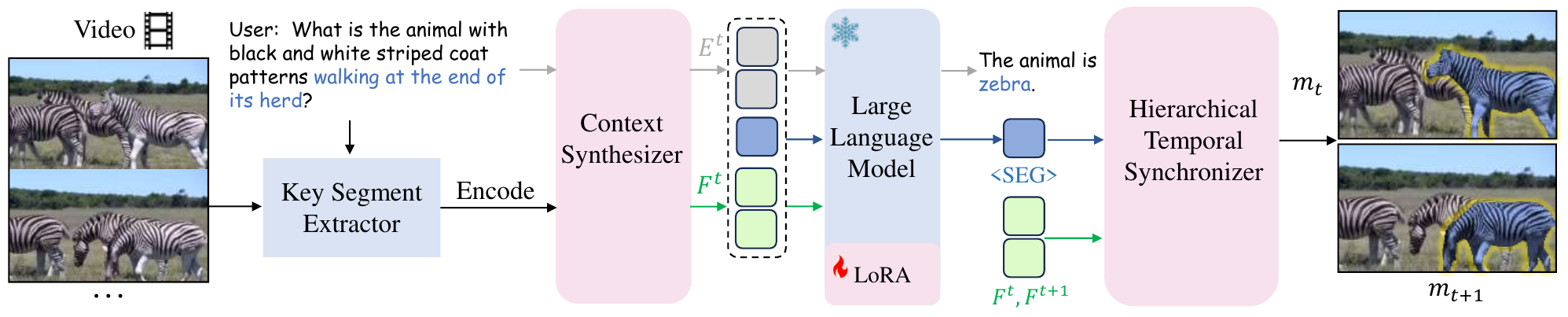}
%Our model employs frame-level queries and video-level ones in the transformer decoder to model object segmentation and tracking respectively.
%\vspace{-5pt}
\caption{{Overall framework of {\modelname}.} Given the input video frames and user input text query, the Key Segment Extractor is proposed to select the most query-relevant video segments from the video. Then, the Context Synthesizer aggregates text-related visual cues from the visual features to the text embeddings and selects the relevant visual features. The large language model generates text output and segmentation tokens with the input of visual features, visually enriched text embeddings, and pre-defined segmentation tokens. Finally, the segmentation tokens are fed to the Hierarchical Temporal Synchronizer to produce the final output segmentation mask tracklets.}
%\vspace{-5pt}
\label{fig:model_graph}
\end{figure*}

%% file: Sec/3_Method.tex
% \section{Video Reasoning Segmentation}
\vspace{-5pt}
\section{\modelname}
% \subsection{Problem Definition}
Video reasoning segmentation aims to output a binary segmentation mask sequence $\{m_t\}_{t=1}^{T} \in \mathbb{R}^{T \times H \times W}$, given an input video clip consisting of $T$ frames $V \in \mathbb{R}^{T \times H \times W \times 3}$ and a query instruction $\mathbf{x}_{txt}$. The task shares a similar formulation with the referring video object segmentation (VOS), but it has complex queries that require reasoning. 

% Apart from the straightforward description, such as ``a zebra to the left'', the query text of video reasoning segmentation includes more complex and nuanced descriptions of the instance (e.g., ``a type of African mammal known for their distinctive black and white striped coat patterns walking at the end of its herd''), involving complicated reasoning as well as the world knowledge. Also, the query text of video reasoning segmentation includes more motion information, such as ``a person wearing a white shirt is in the ocean on a surfboard riding a wave'', which requires the model to capture the moving trajectory of the target. Thus, this topic is challenging and worth exploring, especially in a time when video inputs have become ubiquitous.
\subsection{Model Design}
\noindent\textbf{Model Overview.} As depicted in Fig.~\ref{fig:model_graph}, {\modelname} features a streamlined architecture, comprising mainly several main parts: 1) a pre-trained visual encoder that aligns with a text encoder, 2) a large language model (LLM), 3) a key segment extractor, 4) a context synthesizer, and 5) a hierarchical temporal synchronizer.

The key segment extractor aims to extract the pivotal video segments from a long video input. The context synthesizer and the hierarchical temporal synchronizer are crucial in equipping the LLMs with the capacity to generate masks across temporal dimensions. We utilize the context synthesizer to aggregate the target-text-relevant visual features of the current frame to the input text embeddings. The output text embeddings that carry the most crucial visual cues are then fed into the LLM along with the visual features. In the hierarchical temporal synchronizer, we use the video-scale segmentation tokens to interact with frame-level tokens to produce the final refined complete segmentation tokens that contain both video-level and frame-level information. Finally, the segmentation tokens produce the prediction masks in conjunction with the multi-scale video features. 

\noindent\textbf{Encoder.} For input frame $V_\text{t}$, the visual encoder $\mathcal{I}$ extracts multi-scale visual features $F_{{t}} = \{ F_{{t}}^{\ell} \}_{\ell=1}^{L}$ from $\mathcal{I}(V_{{t}})$, comprising $L$ visual features at the selected layers of $\mathcal{I}$. Meanwhile, the user instruction is fed into the text encoder with the visual embedding $F_t^{L} \in \mathbb{R}^{N \times C}$ (where $N = H/p \times W/p$ and $C$ indicate the number of frame patches and embedding channels, respectively) and generates the $\mathbf{x}_{txt} \in \mathbb{R}^{M \times C}$, where $M$ denotes the number of queries. We choose QFormer as the text encoder to maintain the cross-modality alignment. By incorporating this approach, the text embeddings now capture visual cues that are highly relevant to the user's instructions. The output of the final layer, $F_t^{L}$, encodes global image information and is transformed to align with the language space of LLMs via a vision-to-language projection layer $\text{Proj}_{V\rightarrow L}$. 

% It takes the text embeddings and visual features as input and formulates the  
\vspace{-10pt}
\subsection{Key Segment Extractor} %TemporalSegmentor FrameFocus ChronoSegment
For short videos that typically has very few actions and objects, the segmentation can be solved by image-level reasoning segmentation methods in a frame-by-frame style. These cases, however, can't reflect the temporal reasoning capability of video reasoning segmentation models. We argue that the capability of handling complex videos, which include occluded scenes and longer duration, truly reflects the model's temporal reasoning capacity. As shown in Fig.~\ref{fig:model_graph}, the user input could be answered by segments that contain key information, while the rest of the video is irrelevant. Stemming from this idea, we implement Grounded-VideoLLM~\cite{wang2024grounded}, a SOTA multi-modal LLM capable of grounding the input user queries in long videos, to serve as our Key Segment Extractor. We use a prompt specifying the user query and ask the model to output the starting and ending timestamps of the segments that describe the content of the user query. We take the average from the top-K responses to obtain the starting and ending frames of the key segments, where we denote as $V_{key}$, comprising $T_{key}$ frames. Based on the key segments, we also sample $T_{ref}$ using an adaptive global sampling strategy. We will specify the details of different sampling strategies in the appendix. The extracted frames are then fed into the encoder.
\vspace{-10pt}
\subsection{Context Synthesizer} 
 This module aims to aggregate text-related visual features and inject them to generate text embedding representing the current frame. With the text embeddings $\mathbf{x}_{txt}$ and visual features, context-based aggregation is formulated as: 
\begin{align} \label{equ:refine} 
&   \textbf{E}_{t} = \text{FFN}(\text{CrossAttn}(\mathbf{x}_{txt},F_t^{L},F_t^{L})), \\
&   \textbf{E}_{t}^{c} = \text{Concat} \{ \textbf{E}_{t}^i \}_{i=1}^{K},
\end{align}
where `CrossAttn' refers to cross-attention operation, and $\mathbf{x}_{txt}$ is the query, while $F_t^{L}$ is the value and key. By the cross-attention, we further synthesize the visual cues into the refined text embeddings $\textbf{E}_{t}$. However, we argue that not all queries are needed for the refined text embeddings. Unlike QFormer which adopts 32 queries as input LLM tokens, we propose to condense these embeddings into more compact ones. After getting $\textbf{E}_{t}$, we select the output embeddings with $K$ highest-response scores from the attention matrix out of $M$. Therefore, the final input embeddings $\textbf{E}_{t}^{c}$ preserve the most relevant visuals.

\noindent\textbf{Multi-level Segmentation Tokens.}
To enrich the target-specific encoding and thereby facilitate the generation of high-quality mask tracklets across frames, we devise multi-scale segmentation tokens, representing both frame-level and video-level concepts, tailored to meet the characteristics of video segmentation tasks that demand the modeling of multiple target movements. Formally, we define $C_{\text{seg}} = \left\{ c_n^{s} \in \mathbb{R}^d \right\}_{n=1}^{N}$, where $s\in\{f, v\}$ indicates the frame- or video-scale tokens, $N$ denotes the number of tokens per scale, and $d$ refers to the hidden dimension of LMMs. The multi-level segmentation tokens $C_{\text{seg}}$, combined with the visual features $F_t^{L}$, and condensed text embeddings $\textbf{E}_{t}^{c}$, are then processed by the LLM to generate the response $y_{\text{out}}$:
\begin{align}
y_{\text{out}} = \mathcal{F} (\text{Proj}_{V\rightarrow L}(F_t^{L}), \textbf{E}_{t}^{c}, C_{\text{seg}}).  
\end{align}
Using an example can provide a more vivid explanation of how Large Language Models generate responses. When the user inputs ``segment the leftmost zebra in the video'', the output will include not only a textual answer but also the segmentation tokens. $C_{\text{seg}}$: ``The zebra is $c_1^f, c_1^v$'' (considering the easiest case where $N=1$ for $C_{\text{seg}}$). Before the decoder, an additional projection layer $\Phi$ is utilized to adjust the dimensions of segmentation tokens, in the form of $Q^s = \Phi(C_{\text{seg}})$, ($s\in\{f, v\}$). The output multi-scale segmentation embeddings $Q^v, Q^f$ derived from $C_{\text{seg}}$ are inputs to the pixel decoder $\mathcal{D}$ with $F_{{t}}$ for mask tracklet generation.

\subsection{Hierarchical Temporal Synchronizer}
We design a novel hierarchical temporal synchronizer $\mathcal{D}$ to generate mask tracklets from the multi-scale visual features $F_{{t}}$ and segmentation tokens $Q^v, Q^f$. Derived from the design of the transformer decoder in Mask2Former, $\mathcal{D}$ consists of $L$ decoder layers, where $l^{\text{th}}$ layer cascades a masked cross-attention $h_{\text{CA}}^l$, a self-attention $h_{\text{SA}}^l$, and a feed-forward network $\text{FFN}^l$. Now, we have two scales of segmentation embeddings, $Q^f$ focusing on every frame separately, while $Q^v$ interacting with the whole video features. 
Both the frame- and video-level segmentation embeddings pass the transformer decoder, and are updated as follows:
\begin{align}
\mathbf{Q}^{l+1,s} = \text{FFN}^{\text{s}}(h_{\text{SA}}^{\text{s}}(h_{\text{CA}}^{\text{s}}(\mathbf{Q}^{l,s}, {F_t^l}))),
\end{align}
where $s\in\{f, v\}$, and the $h_{\text{CA}}^{s}(q, r)$ indicates the cross-attention with query embedding $q$ and reference embedding $r$. Then, we feed the frame- and video-level embeddings in aggregation in order to interchange video-level and frame-level information, and this process can be formulated as:
% \begin{align}
%     \mathbf{Q}^{l+1,v^{'}} = \gamma \cdot \text{Softmax}(({Q}^{l+1,v} \times {{Q}^{l+1,f}}^{\top}) \times {{Q}^{l+1,f}}) + (1-\gamma) \cdot \mathbf{Q}^{l+1,v},
% \end{align}
\begin{align}
    \mathbf{Q}^{l+1,v'} &= \gamma \cdot \text{Softmax}\left((\mathbf{Q}^{l+1,v} \times \mathbf{Q}^{l+1,f\top}) \times \mathbf{Q}^{l+1,f}\right) \nonumber\\
                       &\quad + (1-\gamma) \cdot \mathbf{Q}^{l+1,v}
\end{align}
where $\mathbf{Q}^{l+1,v^{'}}$ is the modulated video-level segmentation embeddings at  $l^{\text{th}}$ layer. In this aggregation process, we aggregate the frame-level embeddings with higher responses to our video-level segmentation embeddings in a momentum-based manner. The momentum factor $\gamma$, empirically set to 0.03, governs the update of the video-level embeddings. This choice is based on the assumption that the aggregation of frame-level embeddings should not induce substantial changes to the overall video-level representations.

% \subsection{Training}
\noindent
\textbf{Training Objectives.} The model is trained end-to-end using text generation loss $\mathcal{L}_{txt}$ and segmentation mask loss $\mathcal{L}_{mask}$. The overall objective $\mathcal{L}$ is weighted sum of these losses, determined by $\lambda_{txt}$ and $\lambda_{mask}$: $\mathcal{L} = \lambda_{txt} \mathcal{L}_{txt} + \lambda_{mask} \mathcal{L}_{mask}$, where mask loss $\lambda_{mask}$ is divided into binary cross-entropy loss and dice loss for both video-level and frame-level: $\mathcal{L}_{mask}^s = \lambda_{ce}^s\mathcal{L}_{ce}^s + \lambda_{dice}\mathcal{L}_{dice}$ ($s\in\{f, v\}$), and $\mathcal{L}_{txt}$ is auto-regressive cross-entropy loss for text generation. 

\input{Tab/ReasonVOS}
\input{Tab/refer}

\begin{table}[t]
    \centering
    \caption{\textbf{Video Reasoning Segmentation on ReVOS.}}
    \label{tab:revos}
    \scriptsize
    \vspace{-5pt}
    \resizebox{1.0\linewidth}{!}
    {
        \tablestyle{10pt}{1.0}
        \begin{tabular}{l | c c c}
        \toprule
        \multirow{2}*{Method} & \multicolumn{3}{c}{ReVOS} \\
        &  \( \mathcal{J} \)\&\( \mathcal{F} \) & \( \mathcal{J} \) & \( \mathcal{F} \) \\
        \midrule
        LISA-LLAVA-13B        &41.8  &39.6  &43.9  \\
        VISA-Chat-UniVi-13B   &50.9  &48.8  &52.9  \\
        VISA-InternVideo2-6B  &52.4  &50.1  &54.7  \\
        \cellcolor[HTML]{efefef}\textbf{ViLLa-InternVideo2-6B} &\cellcolor[HTML]{efefef}\textbf{57.0} &\cellcolor[HTML]{efefef}\textbf{54.9} &\cellcolor[HTML]{efefef}\textbf{59.1}\\
        \bottomrule 
        \end{tabular}
    }
\vspace{-14pt}
\end{table}

\subsection{VideoReasonSeg}
For a more comprehensive evaluation of VRS, we establish \textit{VideoReasonSeg}, a dataset containing text instructions, corresponding high-quality masks as well as multiple choices. To ensure reliable assessment, we collect a diverse set of videos from Youtube-VIS~\cite{yang2019video}, OVIS~\cite{qi2022occluded}, LV-VIS~\cite{wang2023LVVIS}, MovieChat~\cite{song2023moviechat}, and VideoMME~\cite{fu2024video}. Videos are categorized based on their duration, including short ($<1$ minutes) and long ($>5$ minutes). To fully evaluate the reasoning capabilities, we design two types of evaluation standards: 1) multiple-choice QAs; 2) instructions and answers. Considering that not all the annotations of selected datasets follow the multiple-choice QA format, we automatically convert the video annotations into this format via LLMs. This is mainly because the open-ended answers have to be scored by LLMs or user studies, which may either introduce evaluation bias or manual intervention. Ultimately, we produce 2 multiple-choice QA pairs for each of the video clips. As for the answers, we mainly focus on the quality of segmentation masks. The resulting \textit{VideoReasonSeg} benchmark comprises a total of 3k videos with 15k object-instruction pairs. All videos are partitioned into two splits: \texttt{train} and \texttt{val}, containing 2k (1820 short videos and 180 long ones) and 1k videos (880 short videos and 120 long ones) respectively. The dataset incorporates three distinct complex scenarios: multiple object segmentation ($\sim$800), rapid motion ($\sim$200), and occlusion in long videos ($\sim$300). The details of data annotation are given in the appendix.

\noindent \textbf{Evaluation Metrics.}
We follow previous works ~\cite{seo2020urvos, ding2023mevis} to adopt $\mathcal{J}\&\mathcal{F}$ as the main evaluation metric, which is the average of region similarity $\mathcal{J}$ and contour accuracy $\mathcal{F}$. Besides, we adopt accuracy as the metric for multiple choices.

\noindent\textbf{Dataset Generation Pipeline.}
We utilize GPT-4V for generating questions and multiple choices. Specifically, we feed all the instance category names and corresponding masks in the video to GPT-4V. With our carefully designed prompts, GPT-4V autonomously selects instances to construct question-answer pairs based on the video contents. Examples of such prompts are illustrated in the appendix.

%% file: Tab/ReasonVOS.tex
\begin{table*}[t]
    %\footnotesize
    %\vspace{0.1cm}
    \centering
    \caption{\textbf{Video Reasoning Segmentation on VideoReasonSeg benchmark.} ``Seg'' refers to ``Segmentation'' while ``MC'' indicates ``Multiple Choices''. ViLLa consistently outperforms other previous methods with different backbones and on different metrics by a large margin. Noted that previous methods are retrained to include our new proposed VideoReasonSeg in the training set.}
    \label{table:reason_seg}
    \vspace{-5pt}
    \resizebox{0.95\linewidth}{!}
    {
        \tablestyle{14pt}{1.0}
        % \begin{footnotesize}
        \begin{tabular}{ l | cc  | cc  | cc   }
            \toprule
            
            \multirow{3}*{Method} & \multicolumn{2}{c|}{Short} & \multicolumn{2}{c|}{Long} & \multicolumn{2}{c}{Overall} \\ 
            
            % \specialrule{0em}{0pt}{1pt}
            \cmidrule{2-7}
            % \cmidrule(lr){1-2} \cmidrule(lr){2-3} \cmidrule(lr){3-4} \cmidrule(lr){4-5} 
            ~ & {Seg} & {MC} & {Seg} & {MC} & {Seg} & {MC} \\
            ~ & \( \mathcal{J} \)\&\( \mathcal{F} \) & Accuracy &\( \mathcal{J} \)\&\( \mathcal{F} \) & Accuracy & \( \mathcal{J} \)\&\( \mathcal{F} \) & Accuracy\\ 
            \midrule
            MTTR~\cite{botach2022end} &17.2 &- &12.7 &- & 15.6 & -\\
            ReferFormer~\cite{wu2022language}  &17.5 &- &13.4 &- & 16.0 & -   \\
            LMPM~\cite{ding2023mevis}  &26.4 &- &18.5 &- &23.1 & - \\
            OnlineRefer~\cite{wu2023onlinerefer}  &26.9 &- &18.7 &- & 16.3 & -\\
            % Mask2Former~\cite{cheng2022masked} & 37.9  & 35.8 & 36.5  & 34.3\\
            % VITA~\cite{heo2022vita} & 39.3  & 36.5 &38.8   &35.7  \\
                % 
            %\midrule
            \midrule
            LISA-LLaVA-7B~\cite{lai2023lisa} &43.8 &34.5 &27.3 &29.5 &38.4 &32.1 \\
            LISA-LLaVA-13B~\cite{lai2023lisa} &45.0 &35.2 &28.1 &30.7 &39.2 &32.9 \\
            % PixelLM-LLaVA-7B~\cite{ren2023pixellm}  &33.2 &42.5 &30.4 &37.2 \\
            % PixelLM-LLaVA-13B~\cite{ren2023pixellm} &34.2 &44.6 &31.8 &41.3 \\
            VideoLISA*-InternVideo2-1\text{B}~\cite{yan2024visa} &51.0 &38.4 &32.4 &32.9 &45.3 &36.7\\
            VideoLISA*-InternVideo2-6\text{B}~\cite{yan2024visa} &54.8 &43.6 &33.8 &34.9 &47.8 &39.2  \\
            VISA-LLaVA-7B~\cite{yan2024visa} &48.9 &37.5 &30.4 &31.9 &43.1 &35.2 \\
            VISA-LLaVA-13B~\cite{yan2024visa} &54.6 &40.1 &33.8 &33.2 &47.4 &37.9 \\
            VISA-Chat-UniVi-13B~\cite{yan2024visa} &54.8 &42.6 &34.0 &34.3 &47.6 &39.6 \\
            VISA-InternVideo2-6\text{B}~\cite{yan2024visa} &57.9 &44.6 &36.2 &35.8 &50.5 &40.5 \\
            {\modelname}-InternVideo2-1\text{B}~\cite{wang2024internvideo2} &57.4 &46.5 &38.9 &37.6 &51.2 &43.1 \\

            \cellcolor[HTML]{efefef}\textbf{{\modelname}-InternVideo2-6\text{B}}~\cite{wang2024internvideo2} &\cellcolor[HTML]{efefef}\textbf{62.6}  &\cellcolor[HTML]{efefef}\textbf{53.8}  &\cellcolor[HTML]{efefef}\textbf{42.0}   &\cellcolor[HTML]{efefef}\textbf{41.0} &\cellcolor[HTML]{efefef}\textbf{55.4} &\cellcolor[HTML]{efefef}\textbf{49.9} \\
            
            % LISA-Llama2-13B (ft) & \textbf{60.0} & \textbf{67.8} & \textbf{43.9} & \textbf{45.8} & \textbf{54.0} & 53.8 & 51.5 & \textbf{51.3} \\
            
            \bottomrule            
        \end{tabular}
        % \end{footnotesize}
    }
\vspace{-5pt}
\end{table*}

%% file: Tab/refer.tex
\begin{table*}[t]
    \centering
    \vspace{-1pt}
    % \caption{\textbf{Comparison on referring video segmentation and video instance segmentation datasets.} }
    \label{tab:recon_parameters}
    
    \centering
    \caption{\textbf{Referring Video Object Segmentation on Ref-YouTube-VOS, Ref-DAVIS17, and MeViS.} ViLLa consistently outperforms other methods on different benchmarks and on different metrics by a large margin.}
    \label{tab:refvos}
    \vspace{-5pt}
    \resizebox{0.95\linewidth}{!}{
        \tablestyle{11pt}{0.9}
        \begin{tabular}{l|ccc|ccc|ccc}
        \toprule
        \multirow{2}*{Method} &  \multicolumn{3}{c |}{Ref-YouTube-VOS} & \multicolumn{3}{c|}{Ref-DAVIS17} &\multicolumn{3}{c}{MeViS}\\
        \cmidrule{2-10}
          & \( \mathcal{J} \)\&\( \mathcal{F} \) & \( \mathcal{J} \) & \( \mathcal{F} \)  &  \( \mathcal{J} \)\&\( \mathcal{F} \) & \( \mathcal{J} \) & \( \mathcal{F} \)   &  \( \mathcal{J} \)\&\( \mathcal{F} \) & \( \mathcal{J} \) & \( \mathcal{F} \) \\
        \midrule
        URVOS~\cite{seo2020urvos}   &47.2 &45.3 &49.2 &51.6  &47.3 &56.0 &27.8 &25.7 &29.9 \\
        LBDT~\cite{ding2022language} &49.4 &48.2 &50.6  &54.1&-&- &29.3 &27.8 &30.8 \\
        ReferFormer~\cite{wu2022language}  & 62.9 & 61.3 & 64.6 & 61.1 & 58.1 & 64.1 & 31.0 & 29.8 & 32.2  \\
        LMPM~\cite{ding2023mevis} &-&-&-&-&-&- &37.2 &34.2 &40.2 \\
        OnlineRefer~\cite{wu2023onlinerefer}  &62.9& 61.0 &64.7 &62.4 &59.1 &65.6 & - & - & -\\
        DsHmp~\cite{he2024decoupling} &67.1 &65.0 &69.1 &64.9 &61.7 &68.1 &46.4 &43.0 &49.8\\
        %\midrule
        \midrule
        LISA~\cite{lai2023lisa}  &54.4 &54.0 &54.8 &66.0 &63.2 &68.8 &37.9 & 35.8 & 40.0\\
        PixelLM~\cite{ren2023pixellm}  &55.0 &54.3 &55.7 &66.7 &63.4 &70.0 &38.7 &36.3 &41.1   \\
        VideoLISA~\cite{bai2024one} &61.7 &60.2 &63.3 &67.7 &63.8 &71.5 &42.3 &39.4 &45.2\\    
        VISA~\cite{yan2024visa} &63.0 &61.4 &64.7 &70.4 &67.0 &73.8 &44.5 &41.8 &47.1\\
        % {\modelname} &66.0 &64.1& 67.9& 64.2 &61.0 &67.4\\
        \cellcolor[HTML]{efefef}\textbf{\modelname} &\cellcolor[HTML]{efefef}\textbf{67.5}& \cellcolor[HTML]{efefef}\textbf{64.6}& \cellcolor[HTML]{efefef}\textbf{70.4}&\cellcolor[HTML]{efefef}\textbf{74.3} &\cellcolor[HTML]{efefef}\textbf{70.6} &\cellcolor[HTML]{efefef}\textbf{78.0} &\cellcolor[HTML]{efefef}\textbf{49.4} &\cellcolor[HTML]{efefef}\textbf{46.5} &\cellcolor[HTML]{efefef}\textbf{52.3}\\
        \bottomrule 
        \end{tabular}
    
    }
    \vspace{-8pt}
\end{table*}

%% file: Sec/5_Experiment.tex
\vspace{-5pt}
\section{Experiment}
In this section, we first present the implementation details and the evaluation benchmark, and then show the comparison results on our proposed reasoning benchmarks and referring segmentation benchmarks. Finally, we ablate on the key components in ViLLa.

\input{Tab/backbone_key_design}
\vspace{-5pt}
\input{Tab/ablation_decoder}

\subsection{Implementation Details}
We use the pre-trained multimodal model from InternVideo2-1B and 6B ~\cite{wang2024internvideo2}, with LoRA for fine-tuning. For the visual encoder and LLM, we apply pre-trained UMT-L from the InternVideo2 stage1 model as the vision encoder and Vicuna-7B v0. Following BLIP2, we deploy QFormer using the pre-trained $BERT_{base}$. To generate multi-scale visual features, we adopt the ViT-Adapter to generate the necessary multi-scale features for the hierarchical temporal synchronizer.  Note that video segmentation differs from image segmentation in that video segmentation involves capturing inter-frame relationships across frames during training. This added complexity results in higher GPU memory requirements compared to image segmentation. To address this challenge, an efficient version of the VIT-Adapter is applied by removing all injectors. 

We use 8 A100 GPUs for training 50 epochs (approximately 1.5 days). We use the AdamW optimizer with the learning rate and weight decay set to $2e-5$ and $0.02$, respectively. The batch size per device is set to 4, the input frame number is 8, the resolution is 224, and the gradient accumulation step is set to 10. Further details are presented in the appendix.

%\vspace{-5pt}
\subsection{Benchmarks}

\noindent
\textbf{Benchmarks.} 
We evaluate ViLLa on benchmarks with video segmentation and question-answering evaluations: VideoReasonSeg, referring video segmentation (Refer-Youtube-VOS, Ref-DAVIS17), ReVOS, and conventional video segmentation (Youtube-VIS series, in the appendix). Through this evaluation, we validate the versatility of ViLLa in diverse segmentation tasks. For the VideoReasonSeg and referring video segmentation benchmark, we use the $\mathcal{J}\&\mathcal{F}$ as well as average accuracy to evaluate the video reasoning of models. In video segmentation (shown in the Appendix), we formulate the queries following the dataset's annotations in the format of ``{\tt Please segment the <description>}'', with <description> corresponding to the descriptions of the instance categories. Training involves random sampling from VideoReasonSeg, VQA data, Video instance segmentation, and Referring video segmentation datasets. The model follows the training protocols of PixelLM~\cite{ren2023pixellm}, pre-training on image datasets for segmentation token initialization.

\vspace{-5pt}
\subsection{Main Results}
\noindent
\textbf{Video Reasoning Segmentation.}
Tab.~\ref{table:reason_seg} compares ViLLa with adapted competing methods on video reasoning segmentation. For the image reasoning segmentation method, to adapt to the video segmentation, we add an extra tracker based on SAM-Track~\cite{cheng2023segment}. Since the original LISA only has one segmentation token, we calculate the $\mathcal{J}\&\mathcal{F}$ between the predicted tracklet and the ground truth.

As shown in Tab.~\ref{table:reason_seg}, our method outperforms the previous image reasoning segmentation LISA by a large margin, demonstrating that our model is capable of accomplishing tasks involving complex reasoning across time. Because video reasoning, different from referring video segmentation and image reasoning segmentation, requires the model to possess \textit{reasoning ability} based on the \textit{whole video}. Thus, without comprehensive knowledge of the whole video, the model can not perform well. What's more, the results also indicate that our method exceeds previous reasoning VOS methods (VideoLISA, VISA) significantly. We argue that this is because of their naive design of sampling strategy and segmentation tokens that fail to model complex trajectories. On the other hand, our model improves the association between video and frame via our specially designed context synthesis strategy and decoder structures, distinguishing our method from previous works.

\begin{figure*}[t]
    \centering
    \includegraphics[width=0.9\textwidth]{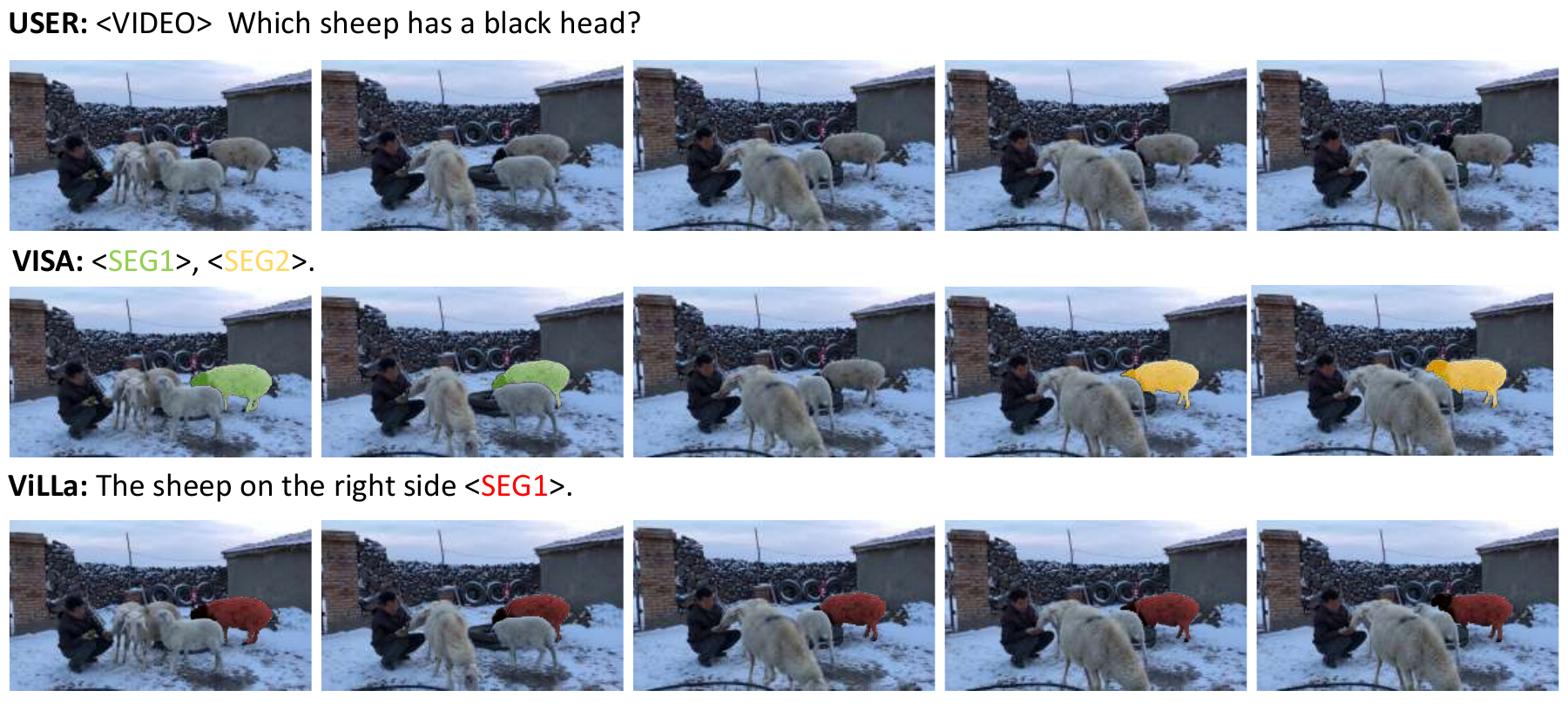}
    \vspace{-10pt}
    \caption{{\textbf{Qualitative Comparisons} between ViLLa and VISA. Compared to VISA, our ViLLa successfully segments the sheep, which demonstrates better discrimination, temporal consistency, and conversation quality.}}
    \label{fig:visualization}
    \vspace{-15pt}
\end{figure*}

\noindent
\textbf{Referring Video Segmentation.}
Tab.~\ref{tab:refvos} presents the results on referring video segmentation datasets. Our ViLLa demonstrates better performances on all these datasets, surpassing VISA by 4.5, 3.9, and 4.9 in $ \mathcal{J} \& \mathcal{F} $ on Ref-Youtube-VOS, Ref-DAVIS17, and MeViS dataset, respectively.

\noindent
\textbf{ReVOS.}
Tab.~\ref{tab:revos} presents the results on ReVOS. ViLLa significantly outperforms previous methods by 4.6 $ \mathcal{J} \& \mathcal{F} $, which further highlights the effectiveness of our design of synthesizer, synchronizer and extractor.
% \noindent
% \textbf{Video Instance Segmentation.}
% Tab.~\ref{tab:vis} presents the results on the video instance segmentation datasets. YouTube-VIS 2019~\cite{yang2019video}, contains 2.9k videos. The dataset was updated to YouTube-VIS 2021 with longer videos. OVIS dataset is another resource for video instance segmentation, particularly focusing on scenarios with severe occlusions between objects~\cite{qi2022occluded}. It consists of 25 object categories and 607 training videos. Our ViLLa surpasses previous SOTA VIS methods by 2.8, 3.4, and 3.6 points, respectively. The results prove that our model is excelling at modeling temporal relations and segmenting high-quality tracklets. 

%\vspace{-5pt}
\subsection{Ablation Studies}
We conduct extensive ablation studies to reveal the contributions of each component. Unless otherwise specified, we conduct the experiments on our VideoReasonSeg dataset.

\noindent 
\textbf{Design Choices of Backbone.}
We emphasize that different vision backbones are applicable in our framework. In Tab.~\ref{tab:ablate_backbone}, we show that InternVideo2 performs the best in both segmentation and multiple choices, potentially because it is the state-of-the-art video foundation model.

\noindent
\textbf{Key Components Design.}
Tab.~\ref{tab:components} demonstrates the effect of our designs based on the backbone and transformer decoder. By adopting our method, we significantly improve the baseline by extracting key information, strengthening the video-frame relationship, and fusing context into text embedding. Tested on VideoReasonSeg, our context synthesizer improves the accuracy of multiple choices by 3.6 points, indicating that synthesizing context to the text embedding and keeping the textual input contextualized is beneficial to video understanding. Also, the design of synchronizer improves the segmentation performance by 1.7 $\mathcal{J}\&\mathcal{F}$, showing that applying multi-scale tokens and synchronizing their relations are helpful for better segmentation and tracking. 

% \noindent
% \textbf{Key Segment Extractor.}

\noindent
\textbf{Hierarchical Temporal Synchronizer.}
Tab.~\ref{tab:decoder} shows the influence of utilizing multi-scale tokens and video-frame synchronization in hierarchical temporal synchronizer. With multi-segment-scale tokens only, the baseline model achieves a 0.5 increase $\mathcal{J}\&\mathcal{F}$, while applying the aggregation brings about an increment of 1.7 $\mathcal{J}\&\mathcal{F}$. This demonstrates that using multi-scale tokens and synchronizing them are beneficial for video segmentation tasks. In Tab.~\ref{tab:scale}, we explore the effects of increasing the number of layers in the decoder. Initially adding layers will contribute to notable gains, but the gain would diminish with more decoder layers.

\vspace{-3pt}
\subsection{Qualitative Results}
\vspace{-2pt}
In Fig.~\ref{fig:visualization}, we provide a visual comparison with VISA. We show that ViLLa is capable of segmenting instances (sheep with a black head) in crowded scenes with heavy occlusion. On the other hand, VISA fails to track the sheep properly after occlusion with other sheep. These results demonstrate the capacity of our ViLLa to handle complicated videos.

%% file: Tab/backbone_key_design.tex
\begin{table*}[t]
    %\vspace{-5pt}
    \centering
    % \caption{\textbf{Comparison on referring video segmentation and video instance segmentation datasets.} }
    \begin{minipage}{0.48\textwidth}
        \centering
        \caption{\textbf{Vision backbones.} InternVideo2 demonstrates better performance (1.5 and 2.4 increase) compared to previous backbones.}
        \vspace{-0.4mm}
        \label{tab:ablate_backbone}   
        \resizebox{0.96\linewidth}{!}
        {
            \tablestyle{10pt}{1.0}
            % \begin{footnotesize}
            \begin{tabular}{ l | c  | c }
                \toprule
                
                \multirow{2}*{Vision Backbone} & {Segmentation} & {Multiple Choices} \\
    
                % \specialrule{0em}{0pt}{1pt}
                % \cline{2-4}
                % \specialrule{0em}{0pt}{1pt}
                
                ~ & \( \mathcal{J} \)\&\( \mathcal{F} \)  & Accuracy\\ 
                
                \midrule
    
                % CLIP-ViT~\cite{} & 26.0 & 14.5 & 17.8 \\
                SAM~\cite{kirillov2023segment} &50.5  & 40.5  \\
                LLaVA-7B~\cite{liu2023llava} & 52.6  &43.1 \\
                InternViT ~\cite{chen2024far} &53.9 &47.5 \\
                \cellcolor[HTML]{efefef}\textbf{InternVideo2}~\cite{wang2024internvideo2} &\cellcolor[HTML]{efefef}\textbf{55.4}   &\cellcolor[HTML]{efefef}\textbf{49.9} \\
                
                % LISA-Llama2-13B (ft) & \textbf{60.0} & \textbf{67.8} & \textbf{43.9} & \textbf{45.8} & \textbf{54.0} & 53.8 & 51.5 & \textbf{51.3} \\
                
                \bottomrule            
            \end{tabular}
            % \end{footnotesize}
        }
    \end{minipage}
    \hfill
    \vspace{10pt}
    \begin{minipage}{0.48\textwidth}
        \centering        
        \caption{\textbf{Key component designs.} Three key modules significantly boost the baseline performance (1.7, 3.6, and 2.2) separately.}
        \label{tab:components}
        \vspace{-8pt}
        \resizebox{\textwidth}{!}{
            \begin{tabular}{l|ccc|cc|c}\toprule
            \multirow{2}{*}{ID}  &Key Segment    &Context        &Hier. Temp.    & \multicolumn{2}{c|}{VidReasonSeg}  & {Refer-VOS} \\ &Extractor &Synthesizer &Synchronizer   &\( \mathcal{J} \)\&\( \mathcal{F} \)  &Accuracy & \( \mathcal{J} \)\&\( \mathcal{F} \) \\
            \midrule
            {1} &   &                     &            &48.5	&39.7	&59.1 \\
            {2} &\checkmark   &           &            &50.2	&43.8	&60.7 \\
            {3} &   &\checkmark           &            &52.1  &43.3	 &63.9 \\
            {4} &   &          &\checkmark             &50.7  &40.9	 &61.5 \\
            {5} &\checkmark   &\checkmark &            &53.7  &48.6  & 65.5    \\
            {6} &\checkmark   &           &\checkmark  &52.2  &44.8    &63.0    \\ 
            {7} &   &\checkmark   &\checkmark          &54.0  &46.0  & 66.2     \\
            \cellcolor[HTML]{efefef}{8} &\cellcolor[HTML]{efefef}\checkmark &\cellcolor[HTML]{efefef}\checkmark &\cellcolor[HTML]{efefef}\checkmark  &\cellcolor[HTML]{efefef}\textbf{55.4} &\cellcolor[HTML]{efefef}\textbf{49.9} &\cellcolor[HTML]{efefef}\textbf{67.5} \\
            
            \bottomrule
            \end{tabular}
        }
    \end{minipage}
    \vspace{-12pt}
\end{table*}

%% file: Tab/ablation_decoder.tex
\begin{table*}[t]
    \centering
    % \caption{\textbf{Comparison on referring video segmentation and video instance segmentation datasets.} }
    \begin{minipage}{0.48\textwidth}
        \centering
        \vspace{-0.2mm}
        \caption{\textbf{Hierarchical Temporal Synchronizer.} Multi-scale tokens and the corresponding synchronization provide significant improvements to the baseline method. Noted that synchronization is based on multi-scale tokens.}
        \label{tab:decoder}
        \vspace{-5pt}
        \resizebox{\linewidth}{!}
        {
            \centering
            \resizebox{\textwidth}{!}{
                \begin{tabular}{l|cc|cc|c}\toprule
                \multirow{2}{*}{ID}      &Multi-scale        &Segment    & \multicolumn{2}{c|}{VidReasonSeg}  & {Refer-VOS} \\ &Tokens &Synchronization   &\( \mathcal{J} \)\&\( \mathcal{F} \)  &Accuracy & \( \mathcal{J} \)\&\( \mathcal{F} \) \\
                \midrule
                {1} &           &            &53.7	&48.6	&65.5 \\
                {2} &\checkmark &            &54.2  &48.9  & 65.8    \\
                \cellcolor[HTML]{efefef}{3} &\cellcolor[HTML]{efefef}\checkmark &\cellcolor[HTML]{efefef}\checkmark  &\cellcolor[HTML]{efefef}\textbf{55.4} &\cellcolor[HTML]{efefef}\textbf{49.9} &\cellcolor[HTML]{efefef}\textbf{67.5} \\
                \bottomrule
                \end{tabular}
            }
            % \end{footnotesize}
        }
    \end{minipage}
    \hfill
    \begin{minipage}{0.48\textwidth}
        \centering
        \caption{\textbf{Hierarchical Temporal Synchronizer.} Adding aggregation layers will contribute to gains, but would diminish gradually.}
        \label{tab:scale}
        %\vspace{-5pt}
        \resizebox{\textwidth}{!}{
            \begin{tabular}{l|cc|cc|c}\toprule
            \multirow{2}{*}{Architecture}       & \multicolumn{2}{c|}{VidReasonSeg}  & {Refer-VOS} \\  &\( \mathcal{J} \)\&\( \mathcal{F} \)  &Accuracy & \( \mathcal{J} \)\&\( \mathcal{F} \) \\
            \midrule
            Baseline            &53.7 &48.6	&65.5 \\
            + 2 layer frame-video aggregation        &54.8  &49.3  & 66.1    \\
            \cellcolor[HTML]{efefef}{+ 3 layer frame-video aggregation}        &\cellcolor[HTML]{efefef}\textbf{55.4} &\cellcolor[HTML]{efefef}\textbf{49.9} &\cellcolor[HTML]{efefef}\textbf{67.5} \\
            + 4 layer frame-video aggregation        &55.1  &49.4  &67.2    \\
            \bottomrule
            \end{tabular}
        }
    \end{minipage}
    \vspace{-10pt}
\end{table*}

%% file: Sec/6_Conclusion.tex
\vspace{-3pt}
\section{Conclusion}
\vspace{-2pt}
We introduce ViLLa, an effective and efficient model that aims to reason and segment in more complex video scenarios. We propose the key segment extractor to select pivotal frames from the video sequence, utilize the context synthesizer to generate text-related visual features, and implement the hierarchical temporal synchronizer to encourage the synchronization between video-level and frame-level segmentation tokens. Also, to help us better tune and evaluate the reasoning segmentation models, we have introduced a benchmark VideoReasonSeg, which comprises 3k video samples with both short and long videos, with both segmentation and QA as evaluations. Benefiting from our exquisite designs, ViLLa demonstrates convincing reasoning video segmentation capabilities in various video scenarios and benchmarks. 

%\noindent\textbf{Limitations and broader impacts.} Even though our model performs well in reasoning segmentation with text instructions, the user input is still limited to text, rather than a variety, such as scribbles. Also, our model can not support multi-round conversations. However, We hope our work can provide new insights into the future direction of combining LLMs and vision tasks so as to build an omnipotent perception model.

%% file: Tab/sampling.tex
\begin{table}[t]
    \centering
    \caption{\textbf{Ablation study} on different sampling strategies.}
    \label{tab:sampling}
    \scriptsize
    \vspace{-5pt}
    \resizebox{1.0\linewidth}{!}
    {
        \tablestyle{20pt}{0.8}
        \begin{tabular}{l | c| c c }
        \toprule
        \multirow{2}*{Strategy} &\multirow{2}*{$T_{ref}$} &  \multicolumn{2}{c}{ReasonVideoSeg} \\
         & & \( \mathcal{J} \)\&\( \mathcal{F} \) &Accuracy   \\
        \midrule
        \multirow{3}*{Global} &0 &54.0 &47.8  \\
        &6 &54.3  &48.3  \\
        &12 &54.5  &48.7   \\
        \midrule
        \multirow{3}*{Neighbor} &0 &54.0 &47.8  \\
        &6 &54.4  &48.5    \\
        &12 &54.7   &49.2  \\
        \midrule
        \multirow{3}*{\textbf{Adaptive}} &0 &54.0 &47.8 \\
        &6 &54.8 &49.0\\ 
        &12 &\textbf{55.4}& \textbf{49.9}\\
        \bottomrule 
        \end{tabular}
    }
\vspace{-5pt}
\end{table}

%% file: main.bbl
\begin{thebibliography}{10}

\bibitem{alayrac2022flamingo}
Jean-Baptiste Alayrac, Jeff Donahue, Pauline Luc, Antoine Miech, Iain Barr, Yana Hasson, Karel Lenc, Arthur Mensch, Katherine Millican, Malcolm Reynolds, et~al.
\newblock Flamingo: a visual language model for few-shot learning.
\newblock In {\em NeurIPS}, 2022.

\bibitem{athar2020stem}
Ali Athar, Sabarinath Mahadevan, Aljosa Osep, Laura Leal-Taix{\'e}, and Bastian Leibe.
\newblock Stem-seg: Spatio-temporal embeddings for instance segmentation in videos.
\newblock In {\em ECCV}, 2020.

\bibitem{Qwen-VL}
Jinze Bai, Shuai Bai, Shusheng Yang, Shijie Wang, Sinan Tan, Peng Wang, Junyang Lin, Chang Zhou, and Jingren Zhou.
\newblock Qwen-vl: A versatile vision-language model for understanding, localization, text reading, and beyond.
\newblock {\em arXiv:2308.12966}, 2023.

\bibitem{bai2024one}
Zechen Bai, Tong He, Haiyang Mei, Pichao Wang, Ziteng Gao, Joya Chen, Lei Liu, Zheng Zhang, and Mike~Zheng Shou.
\newblock One token to seg them all: Language instructed reasoning segmentation in videos.
\newblock In {\em NeurIPS}, 2024.

\bibitem{botach2022end}
Adam Botach, Evgenii Zheltonozhskii, and Chaim Baskin.
\newblock End-to-end referring video object segmentation with multimodal transformers.
\newblock In {\em CVPR}, 2022.

\bibitem{chen2024far}
Zhe Chen, Weiyun Wang, Hao Tian, Shenglong Ye, Zhangwei Gao, Erfei Cui, Wenwen Tong, Kongzhi Hu, Jiapeng Luo, Zheng Ma, et~al.
\newblock How far are we to gpt-4v? closing the gap to commercial multimodal models with open-source suites.
\newblock {\em arXiv:2404.16821}, 2024.

\bibitem{cheng2021mask2former}
Bowen Cheng, Anwesa Choudhuri, Ishan Misra, Alexander Kirillov, Rohit Girdhar, and Alexander~G Schwing.
\newblock Mask2former for video instance segmentation.
\newblock {\em arXiv:2112.10764}, 2021.

\bibitem{cheng2022masked}
Bowen Cheng, Ishan Misra, Alexander~G Schwing, Alexander Kirillov, and Rohit Girdhar.
\newblock Masked-attention mask transformer for universal image segmentation.
\newblock In {\em CVPR}, 2022.

\bibitem{cheng2022xmem}
Ho~Kei Cheng and Alexander~G Schwing.
\newblock Xmem: Long-term video object segmentation with an atkinson-shiffrin memory model.
\newblock In {\em ECCV}, 2022.

\bibitem{cheng2023segment}
Yangming Cheng, Liulei Li, Yuanyou Xu, Xiaodi Li, Zongxin Yang, Wenguan Wang, and Yi~Yang.
\newblock Segment and track anything.
\newblock {\em arXiv:2305.06558}, 2023.

\bibitem{vicuna}
Wei-Lin Chiang, Zhuohan Li, Zi~Lin, Ying Sheng, Zhanghao Wu, Hao Zhang, Lianmin Zheng, Siyuan Zhuang, Yonghao Zhuang, Joseph~E Gonzalez, et~al.
\newblock Vicuna: An open-source chatbot impressing gpt-4 with 90\% chatgpt quality, 2023.

\bibitem{instructblip}
Wenliang Dai, Junnan Li, Dongxu Li, Anthony Meng~Huat Tiong, Junqi Zhao, Weisheng Wang, Boyang Li, Pascale Fung, and Steven Hoi.
\newblock Instructblip: Towards general-purpose vision-language models with instruction tuning.
\newblock {\em arXiv:2305.06500}, 2023.

\bibitem{ding2023mevis}
Henghui Ding, Chang Liu, Shuting He, Xudong Jiang, and Chen~Change Loy.
\newblock Mevis: A large-scale benchmark for video segmentation with motion expressions.
\newblock In {\em ICCV}, 2023.

\bibitem{ding2022language}
Zihan Ding, Tianrui Hui, Junshi Huang, Xiaoming Wei, Jizhong Han, and Si~Liu.
\newblock Language-bridged spatial-temporal interaction for referring video object segmentation.
\newblock In {\em CVPR}, 2022.

\bibitem{hao2024vitron}
Hao Fei, Shengqiong Wu, Hanwang Zhang, Tat-Seng Chua, and Shuicheng Yan.
\newblock Vitron: A unified pixel-level vision llm for understanding, generating, segmenting, editing.
\newblock In {\em NeurIPS}, 2024.

\bibitem{fu2024video}
Chaoyou Fu, Yuhan Dai, Yondong Luo, Lei Li, Shuhuai Ren, Renrui Zhang, Zihan Wang, Chenyu Zhou, Yunhang Shen, Mengdan Zhang, et~al.
\newblock Video-mme: The first-ever comprehensive evaluation benchmark of multi-modal llms in video analysis.
\newblock In {\em CVPR}, 2025.

\bibitem{llama-adapter-v2}
Peng Gao, Jiaming Han, Renrui Zhang, Ziyi Lin, Shijie Geng, Aojun Zhou, Wei Zhang, Pan Lu, Conghui He, Xiangyu Yue, et~al.
\newblock Llama-adapter v2: Parameter-efficient visual instruction model.
\newblock {\em arXiv:2304.15010}, 2023.

\bibitem{he2020momentum}
Kaiming He, Haoqi Fan, Yuxin Wu, Saining Xie, and Ross Girshick.
\newblock Momentum contrast for unsupervised visual representation learning.
\newblock In {\em CVPR}, 2020.

\bibitem{he2024decoupling}
Shuting He and Henghui Ding.
\newblock Decoupling static and hierarchical motion perception for referring video segmentation.
\newblock In {\em CVPR}, 2024.

\bibitem{heo2022generalized}
Miran Heo, Sukjun Hwang, Jeongseok Hyun, Hanjung Kim, Seoung~Wug Oh, Joon-Young Lee, and Seon~Joo Kim.
\newblock A generalized framework for video instance segmentation.
\newblock In {\em CVPR}, 2023.

\bibitem{heo2022vita}
Miran Heo, Sukjun Hwang, Seoung~Wug Oh, Joon-Young Lee, and Seon~Joo Kim.
\newblock Vita: Video instance segmentation via object token association.
\newblock In {\em NeurIPS}, 2022.

\bibitem{khoreva2019video}
Anna Khoreva, Anna Rohrbach, and Bernt Schiele.
\newblock Video object segmentation with language referring expressions.
\newblock In {\em ACCV}, 2019.

\bibitem{kirillov2023segment}
Alexander Kirillov, Eric Mintun, Nikhila Ravi, Hanzi Mao, Chloe Rolland, Laura Gustafson, Tete Xiao, Spencer Whitehead, Alexander~C Berg, Wan-Yen Lo, et~al.
\newblock Segment anything.
\newblock In {\em ICCV}, 2023.

\bibitem{lai2023lisa}
Xin Lai, Zhuotao Tian, Yukang Chen, Yanwei Li, Yuhui Yuan, Shu Liu, and Jiaya Jia.
\newblock Lisa: Reasoning segmentation via large language model.
\newblock In {\em CVPR}, 2024.

\bibitem{li2023otter}
Bo~Li, Yuanhan Zhang, Liangyu Chen, Jinghao Wang, Jingkang Yang, and Ziwei Liu.
\newblock Otter: A multi-modal model with in-context instruction tuning.
\newblock {\em arXiv:2305.03726}, 2023.

\bibitem{li2023blip}
Junnan Li, Dongxu Li, Silvio Savarese, and Steven Hoi.
\newblock Blip-2: Bootstrapping language-image pre-training with frozen image encoders and large language models.
\newblock In {\em ICML}, 2023.

\bibitem{OMGSeg}
Xiangtai Li, Haobo Yuan, Wei Li, Henghui Ding, Size Wu, Wenwei Zhang, Yining Li, Kai Chen, and Chen~Change Loy.
\newblock Omg-seg: Is one model good enough for all segmentation?
\newblock In {\em CVPR}, 2024.

\bibitem{li2022videoknet}
Xiangtai Li, Wenwei Zhang, Jiangmiao Pang, Kai Chen, Guangliang Cheng, Yunhai Tong, and Chen~Change Loy.
\newblock Video k-net: A simple, strong, and unified baseline for video segmentation.
\newblock In {\em CVPR}, 2022.

\bibitem{li2025llama}
Yanwei Li, Chengyao Wang, and Jiaya Jia.
\newblock Llama-vid: An image is worth 2 tokens in large language models.
\newblock In {\em ECCV}, 2024.

\bibitem{lin2021video}
Huaijia Lin, Ruizheng Wu, Shu Liu, Jiangbo Lu, and Jiaya Jia.
\newblock Video instance segmentation with a propose-reduce paradigm.
\newblock In {\em ICCV}, 2021.

\bibitem{liu2024referring}
Chang Liu, Xiangtai Li, and Henghui Ding.
\newblock Referring image editing: Object-level image editing via referring expressions.
\newblock In {\em CVPR}, 2024.

\bibitem{liu2023improved}
Haotian Liu, Chunyuan Li, Yuheng Li, and Yong~Jae Lee.
\newblock Improved baselines with visual instruction tuning.
\newblock In {\em CVPR}, 2024.

\bibitem{liu2023llava}
Haotian Liu, Chunyuan Li, Qingyang Wu, and Yong~Jae Lee.
\newblock Visual instruction tuning.
\newblock In {\em NeurIPS}, 2023.

\bibitem{liu2023llavaplus}
Shilong Liu, Hao Cheng, Haotian Liu, Hao Zhang, Feng Li, Tianhe Ren, Xueyan Zou, Jianwei Yang, Hang Su, Jun Zhu, et~al.
\newblock Llava-plus: Learning to use tools for creating multimodal agents.
\newblock In {\em ECCV}, 2024.

\bibitem{2023interngpt}
Zhaoyang Liu, Yinan He, Wenhai Wang, Weiyun Wang, Yi~Wang, Shoufa Chen, Qinglong Zhang, Zeqiang Lai, Yang Yang, Qingyun Li, Jiashuo Yu, et~al.
\newblock Interngpt: Solving vision-centric tasks by interacting with chatgpt beyond language.
\newblock {\em arXiv:2305.05662}, 2023.

\bibitem{lu2022unified}
Jiasen Lu, Christopher Clark, Rowan Zellers, Roozbeh Mottaghi, and Aniruddha Kembhavi.
\newblock Unified-io: A unified model for vision, language, and multi-modal tasks.
\newblock In {\em ICLR}, 2022.

\bibitem{luo2023valley}
Ruipu Luo, Ziwang Zhao, Min Yang, Junwei Dong, Minghui Qiu, Pengcheng Lu, Tao Wang, and Zhongyu Wei.
\newblock Valley: Video assistant with large language model enhanced ability.
\newblock {\em arXiv:2306.07207}, 2023.

\bibitem{maaz2023video}
Muhammad Maaz, Hanoona Rasheed, Salman Khan, and Fahad~Shahbaz Khan.
\newblock Video-chatgpt: Towards detailed video understanding via large vision and language models.
\newblock {\em arXiv:2306.05424}, 2023.

\bibitem{chatgpt}
OpenAI.
\newblock Chatgpt: A language model for conversational ai.
\newblock Technical report, OpenAI, 2023.

\bibitem{openai2023gpt4}
OpenAI.
\newblock Gpt-4 technical report, 2023.

\bibitem{peng2023instruction}
Baolin Peng, Chunyuan Li, Pengcheng He, Michel Galley, and Jianfeng Gao.
\newblock Instruction tuning with gpt-4.
\newblock {\em arXiv:2304.03277}, 2023.

\bibitem{peng2023kosmos}
Zhiliang Peng, Wenhui Wang, Li~Dong, Yaru Hao, Shaohan Huang, Shuming Ma, and Furu Wei.
\newblock Kosmos-2: Grounding multimodal large language models to the world.
\newblock {\em arXiv:2306.14824}, 2023.

\bibitem{qi2022occluded}
Jiyang Qi, Yan Gao, Yao Hu, Xinggang Wang, Xiaoyu Liu, Xiang Bai, Serge Belongie, Alan Yuille, Philip~HS Torr, and Song Bai.
\newblock Occluded video instance segmentation: A benchmark.
\newblock {\em IJCV}, 2022.

\bibitem{qi2024generalizable}
Lu~Qi, Yi-Wen Chen, Lehan Yang, Tiancheng Shen, Xiangtai Li, Weidong Guo, Yu~Xu, and Ming-Hsuan Yang.
\newblock Generalizable entity grounding via assistance of large language model.
\newblock {\em arXiv:2402.02555}, 2024.

\bibitem{rasheed2024glamm}
Hanoona Rasheed, Muhammad Maaz, Sahal Shaji, Abdelrahman Shaker, Salman Khan, Hisham Cholakkal, Rao~M Anwer, Eric Xing, Ming-Hsuan Yang, and Fahad~S Khan.
\newblock Glamm: Pixel grounding large multimodal model.
\newblock In {\em CVPR}, 2024.

\bibitem{ren2023pixellm}
Zhongwei Ren, Zhicheng Huang, Yunchao Wei, Yao Zhao, Dongmei Fu, Jiashi Feng, and Xiaojie Jin.
\newblock Pixellm: Pixel reasoning with large multimodal model.
\newblock In {\em CVPR}, 2024.

\bibitem{sadeghzadeh2024arva}
Arezoo Sadeghzadeh, Md~Baharul Islam, Md~Nur Uddin, and Tarkan Aydin.
\newblock Arva: An augmented reality-based visual aid for mobility enhancement through real-time video stream transformation.
\newblock {\em IEEE ACCESS}, 2024.

\bibitem{seo2020urvos}
Seonguk Seo, Joon-Young Lee, and Bohyung Han.
\newblock Urvos: Unified referring video object segmentation network with a large-scale benchmark.
\newblock In {\em ECCV}, 2020.

\bibitem{sermanet2024robovqa}
Pierre Sermanet, Tianli Ding, Jeffrey Zhao, Fei Xia, Debidatta Dwibedi, Keerthana Gopalakrishnan, Christine Chan, Gabriel Dulac-Arnold, Sharath Maddineni, Nikhil~J Joshi, et~al.
\newblock Robovqa: Multimodal long-horizon reasoning for robotics.
\newblock In {\em ICRA}, 2024.

\bibitem{song2023moviechat}
Enxin Song, Wenhao Chai, Guanhong Wang, Yucheng Zhang, Haoyang Zhou, Feiyang Wu, Xun Guo, Tian Ye, Yan Lu, Jenq-Neng Hwang, et~al.
\newblock Moviechat: From dense token to sparse memory for long video understanding.
\newblock {\em arXiv:2307.16449}, 2023.

\bibitem{alpaca}
Rohan Taori, Ishaan Gulrajani, Tianyi Zhang, Yann Dubois, Xuechen Li, Carlos Guestrin, Percy Liang, and Tatsunori~B Hashimoto.
\newblock Stanford alpaca: An instruction-following llama model, 2023.

\bibitem{touvron2023llama}
Hugo Touvron, Thibaut Lavril, Gautier Izacard, Xavier Martinet, Marie-Anne Lachaux, Timoth{\'e}e Lacroix, Baptiste Rozi{\`e}re, Naman Goyal, Eric Hambro, Faisal Azhar, et~al.
\newblock Llama: Open and efficient foundation language models.
\newblock {\em arXiv:2302.13971}, 2023.

\bibitem{wang2024grounded}
Haibo Wang, Zhiyang Xu, Yu~Cheng, Shizhe Diao, Yufan Zhou, Yixin Cao, Qifan Wang, Weifeng Ge, and Lifu Huang.
\newblock Grounded-videollm: Sharpening fine-grained temporal grounding in video large language models.
\newblock {\em arXiv:2410.03290}, 2024.

\bibitem{wang2020context}
Hao Wang, Cheng Deng, Fan Ma, and Yi~Yang.
\newblock Context modulated dynamic networks for actor and action video segmentation with language queries.
\newblock In {\em AAAI}, 2020.

\bibitem{wang2019asymmetric}
Hao Wang, Cheng Deng, Junchi Yan, and Dacheng Tao.
\newblock Asymmetric cross-guided attention network for actor and action video segmentation from natural language query.
\newblock In {\em ICCV}, 2019.

\bibitem{wang2023LVVIS}
Haochen Wang, Cilin Yan, Shuai Wang, Xiaolong Jiang, Xu~Tang, Yao Hu, Weidi Xie, and Efstratios Gavves.
\newblock Towards open-vocabulary video instance segmentation.
\newblock In {\em ICCV}, 2023.

\bibitem{wang2022ofa}
Peng Wang, An~Yang, Rui Men, Junyang Lin, Shuai Bai, Zhikang Li, Jianxin Ma, Chang Zhou, Jingren Zhou, and Hongxia Yang.
\newblock Ofa: Unifying architectures, tasks, and modalities through a simple sequence-to-sequence learning framework.
\newblock In {\em ICML}, 2022.

\bibitem{wang2024omnidrive}
Shihao Wang, Zhiding Yu, Xiaohui Jiang, Shiyi Lan, Min Shi, Nadine Chang, Jan Kautz, Ying Li, and Jose~M Alvarez.
\newblock Omnidrive: A holistic llm-agent framework for autonomous driving with 3d perception, reasoning and planning.
\newblock {\em arXiv:2405.01533}, 2024.

\bibitem{wang2023internvid}
Yi~Wang, Yinan He, Yizhuo Li, Kunchang Li, Jiashuo Yu, Xin Ma, Xinyuan Chen, Yaohui Wang, Ping Luo, Ziwei Liu, et~al.
\newblock Internvid: A large-scale video-text dataset for multimodal understanding and generation.
\newblock {\em arXiv:2307.06942}, 2023.

\bibitem{wang2024internvideo2}
Yi~Wang, Kunchang Li, Xinhao Li, Jiashuo Yu, Yinan He, Guo Chen, Baoqi Pei, Rongkun Zheng, Jilan Xu, Zun Wang, et~al.
\newblock Internvideo2: Scaling video foundation models for multimodal video understanding.
\newblock {\em arXiv:2403.15377}, 2024.

\bibitem{wang2021end}
Yuqing Wang, Zhaoliang Xu, Xinlong Wang, Chunhua Shen, Baoshan Cheng, Hao Shen, and Huaxia Xia.
\newblock End-to-end video instance segmentation with transformers.
\newblock In {\em CVPR}, 2021.

\bibitem{wu2022multi}
Dongming Wu, Xingping Dong, Ling Shao, and Jianbing Shen.
\newblock Multi-level representation learning with semantic alignment for referring video object segmentation.
\newblock In {\em CVPR}, 2022.

\bibitem{wu2023onlinerefer}
Dongming Wu, Tiancai Wang, Yuang Zhang, Xiangyu Zhang, and Jianbing Shen.
\newblock Onlinerefer: A simple online baseline for referring video object segmentation.
\newblock In {\em ICCV}, 2023.

\bibitem{wu2022efficient}
Jialian Wu, Sudhir Yarram, Hui Liang, Tian Lan, Junsong Yuan, Jayan Eledath, and Gerard Medioni.
\newblock Efficient video instance segmentation via tracklet query and proposal.
\newblock In {\em CVPR}, 2022.

\bibitem{wu2022language}
Jiannan Wu, Yi~Jiang, Peize Sun, Zehuan Yuan, and Ping Luo.
\newblock Language as queries for referring video object segmentation.
\newblock In {\em CVPR}, 2022.

\bibitem{wu2022seqformer}
Junfeng Wu, Yi~Jiang, Song Bai, Wenqing Zhang, and Xiang Bai.
\newblock Seqformer: Sequential transformer for video instance segmentation.
\newblock In {\em ECCV}, 2022.

\bibitem{wu2022defense}
Junfeng Wu, Qihao Liu, Yi~Jiang, Song Bai, Alan Yuille, and Xiang Bai.
\newblock In defense of online models for video instance segmentation.
\newblock In {\em ECCV}, 2022.

\bibitem{yan2024visa}
Cilin Yan, Haochen Wang, Shilin Yan, Xiaolong Jiang, Yao Hu, Guoliang Kang, Weidi Xie, and Efstratios Gavves.
\newblock Visa: Reasoning video object segmentation via large language models.
\newblock {\em arXiv:2407.11325}, 2024.

\bibitem{yan2024referred}
Shilin Yan, Renrui Zhang, Ziyu Guo, Wenchao Chen, Wei Zhang, Hongyang Li, Yu~Qiao, Hao Dong, Zhongjiang He, and Peng Gao.
\newblock Referred by multi-modality: A unified temporal transformer for video object segmentation.
\newblock In {\em AAAI}, 2024.

\bibitem{yang2019video}
Linjie Yang, Yuchen Fan, and Ning Xu.
\newblock Video instance segmentation.
\newblock In {\em ICCV}, 2019.

\bibitem{yang2023improved}
Senqiao Yang, Tianyuan Qu, Xin Lai, Zhuotao Tian, Bohao Peng, Shu Liu, and Jiaya Jia.
\newblock An improved baseline for reasoning segmentation with large language model.
\newblock {\em arXiv:2312.17240}, 2023.

\bibitem{ye2023mplug}
Qinghao Ye, Haiyang Xu, Guohai Xu, Jiabo Ye, Ming Yan, Yiyang Zhou, Junyang Wang, Anwen Hu, Pengcheng Shi, Yaya Shi, et~al.
\newblock mplug-owl: Modularization empowers large language models with multimodality.
\newblock {\em arXiv:2304.14178}, 2023.

\bibitem{ying2023ctvis}
Kaining Ying, Qing Zhong, Weian Mao, Zhenhua Wang, Hao Chen, Lin~Yuanbo Wu, Yifan Liu, Chengxiang Fan, Yunzhi Zhuge, and Chunhua Shen.
\newblock Ctvis: Consistent training for online video instance segmentation.
\newblock In {\em ICCV}, 2023.

\bibitem{you2023ferret}
Haoxuan You, Haotian Zhang, Zhe Gan, Xianzhi Du, Bowen Zhang, Zirui Wang, Liangliang Cao, Shih-Fu Chang, and Yinfei Yang.
\newblock Ferret: Refer and ground anything anywhere at any granularity.
\newblock {\em arXiv:2310.07704}, 2023.

\bibitem{yu2022coca}
Jiahui Yu, Zirui Wang, Vijay Vasudevan, Legg Yeung, Mojtaba Seyedhosseini, and Yonghui Wu.
\newblock Coca: Contrastive captioners are image-text foundation models.
\newblock {\em arXiv:2205.01917}, 2022.

\bibitem{zhang2023llama}
Renrui Zhang, Jiaming Han, Aojun Zhou, Xiangfei Hu, Shilin Yan, Pan Lu, Hongsheng Li, Peng Gao, and Yu~Qiao.
\newblock Llama-adapter: Efficient fine-tuning of language models with zero-init attention.
\newblock {\em arXiv:2303.16199}, 2023.

\bibitem{zhang2022opt}
Susan Zhang, Stephen Roller, Naman Goyal, Mikel Artetxe, Moya Chen, Shuohui Chen, Christopher Dewan, Mona Diab, Xian Li, Xi~Victoria Lin, et~al.
\newblock Opt: Open pre-trained transformer language models.
\newblock {\em arXiv:2205.01068}, 2022.

\bibitem{zhang2024omg}
Tao Zhang, Xiangtai Li, Hao Fei, Haobo Yuan, Shengqiong Wu, Shunping Ji, Chen~Change Loy, and Shuicheng Yan.
\newblock Omg-llava: Bridging image-level, object-level, pixel-level reasoning and understanding.
\newblock {\em arXiv:2406.19389}, 2024.

\bibitem{zhang2023dvis}
Tao Zhang, Xingye Tian, Yu~Wu, Shunping Ji, Xuebo Wang, Yuan Zhang, and Pengfei Wan.
\newblock Dvis: Decoupled video instance segmentation framework.
\newblock {\em arXiv:2306.03413}, 2023.

\bibitem{dvisdaq}
Yikang Zhou, Tao Zhang, Shunping Ji, Shuicheng Yan, and Xiangtai Li.
\newblock Dvis-daq: Improving video segmentation via dynamic anchor queries.
\newblock In {\em ECCV}, 2024.

\bibitem{zhu2023minigpt}
Deyao Zhu, Jun Chen, Xiaoqian Shen, Xiang Li, and Mohamed Elhoseiny.
\newblock Minigpt-4: Enhancing vision-language understanding with advanced large language models.
\newblock {\em arXiv:2304.10592}, 2023.

\bibitem{zhu2020deformable}
Xizhou Zhu, Weijie Su, Lewei Lu, Bin Li, Xiaogang Wang, and Jifeng Dai.
\newblock Deformable detr: Deformable transformers for end-to-end object detection.
\newblock In {\em ICLR}, 2020.

\end{thebibliography}
